\documentclass[10pt,journal,compsoc]{IEEEtran}
\ifCLASSOPTIONcompsoc
  \usepackage[nocompress]{cite}
\else
  \usepackage{cite}
\fi
\ifCLASSINFOpdf
\else
\fi

\usepackage{booktabs} 
\usepackage{xparse}

\usepackage{hyperref}
\usepackage{enumitem}
\usepackage{epsfig}
\usepackage[lofdepth,lotdepth]{subfig}

\usepackage[numbers,square]{natbib}

\usepackage{url}
\usepackage[noend]{algorithmic}
\usepackage{xcolor}
\usepackage{amsmath, amssymb}

\usepackage[ruled]{algorithm}
\usepackage[export]{adjustbox}

\usepackage{pstricks}
\usepackage{pst-all}
\usepackage{pst-plot}
\usepackage{pst-func,pst-math}
\usepackage{pgfplots}
\usepackage{tikz}
\usepackage{etex}
\usepackage{bbm}

\newgray{shadecolor}{0.85}
\definecolor{shadecolor}{gray}{0.85}

\newgray{graylight}{0.75}
\newgray{grayplus}{0.35}
\newgray{graydark}{0.1}

\newcommand{\MovieL}{\textsc{MovieLens}}
\newcommand{\NetF}{\textsc{Netflix}}

\newcommand{\ML}{\textsc{ML}}

\newcommand{\userS}{\mathcal{U}}
\newcommand{\itemS}{\mathcal{I}}
\newcommand{\vecU}{\mathbf{U}}
\newcommand{\vecI}{\mathbf{V}}

\newcommand{\RecNet}{\texttt{RecNet}}

\newcommand{\BPR}{\texttt{BPR-MF}}
\newcommand{\CoFactor}{\texttt{Co-Factor}}
\newcommand{\LightFM}{\texttt{LightFM}}

\newcommand{\Loss}{\mathcal{L}}
\newcommand{\Trn}{\mathcal{S}}

\newcommand{\EE}{\mathbb E}
\newcommand{\Ind}{\mathbbm{1}}

\newcommand{\N}{\mathbb N}
\newcommand{\Input}{\mathcal X}
\newcommand{\R}{\mathbb R}
\newcommand{\prefu}{\renewcommand\arraystretch{.2} \begin{array}{c}
   {\succ} \\  \mbox{{\tiny {\it u}}}
  \end{array}\renewcommand\arraystretch{1ex}}

\newcommand{\graph}{\Omega}

\newcommand{\vertices}{\mathcal V}
\newcommand{\edges}{\mathcal E}
\newcommand{\Cset}{\mathcal M}
\newcommand{\Weight}{W}
\newcommand{\cover}{\mathcal C}

\newcommand{\covers}{{\mathcal K}}
\newcommand{\bfZ}{\mathbf{z}}
\newcommand{\rademacher}{\mathfrak{R}}
\newcommand{\DA}{^\downarrow}
\newcommand{\kasandr}{\textsc{Kasandr}}
\newcommand{\cmmnt}[1]{}

\newtheorem{theorem}{Theorem}
\newtheorem{definition}{Definition}

\usepackage{mathtools}

\begin{document}
\begin{sloppypar}
%
\title{Representation Learning and Pairwise Ranking for Implicit Feedback in Recommendation Systems}
%
%
%
%

\author{Sumit Sidana, Mikhail Trofimov,  Oleg Horodnitskii,  Charlotte Laclau, Yury Maximov, Massih-Reza Amini
\IEEEcompsocitemizethanks{\IEEEcompsocthanksitem Univ. Grenoble Alpes/CNRS.\protect\\
		E-mail: \{fname.lname\}@univ-grenoble-alpes.fr
		\IEEEcompsocthanksitem Federal Research Center "Computer Science and Control" of Russian Academy of Sciences . \protect\\
		E-mail: mikhail.trofimov@phystech.edu
			\IEEEcompsocthanksitem Center for Energy Systems, Skolkovo Institute of Science . \protect\\
		E-mail: Oleg.Gorodnitskii@skoltech.ru
			\IEEEcompsocthanksitem T-4 and CNLS, Los Alamos National Laboratory, and Center for Energy Systems, Skolkovo Institute of Science and Technology . \protect\\
		E-mail: yury@lanl.gov}
	\thanks{Manuscript received April 19, 2005; revised August 26, 2015.}}

\markboth{Journal of \LaTeX\ Class Files,~Vol.~14, No.~8, August~2015}%
{Shell \MakeLowercase{\textit{et al.}}: Bare Demo of IEEEtran.cls for Computer Society Journals}
%



\IEEEtitleabstractindextext{%
\begin{abstract}
In this paper, we propose a novel ranking framework for collaborative filtering with the overall aim of learning user preferences over items by minimizing a pairwise ranking loss. We show the minimization problem involves dependent random variables and provide a theoretical analysis by proving the consistency of the empirical risk minimization in the worst case where all users choose a minimal number of positive and negative items. We further derive a Neural-Network model that jointly learns a new representation of users and items in an embedded space as well as the preference relation of users over the pairs of items. The learning objective is based on three scenarios of ranking losses that control the ability of the model to maintain the ordering over the items induced from the users' preferences, as well as, the capacity of the dot-product defined in the learned embedded space to produce the ordering. The proposed model is by nature suitable for implicit feedback and involves the estimation of only very few parameters. Through extensive experiments on several real-world benchmarks on implicit data, we show the interest of learning the preference and the embedding simultaneously when compared to learning those separately. We also demonstrate that our approach is very competitive with the best state-of-the-art collaborative filtering techniques proposed for implicit feedback.
\end{abstract}

}

\maketitle

\IEEEdisplaynontitleabstractindextext

%
\IEEEpeerreviewmaketitle

\ifCLASSOPTIONcaptionsoff
  \newpage
\fi



\section{Introduction}

In the recent years, recommender systems (RS) have attracted a lot of interest in both industry and academic research communities, mainly due to new challenges that the design of a decisive and efficient RS presents. Given a set of customers (or users), the goal of RS is to provide a personalized recommendation of products to users which would likely to be of their interest. Common examples of applications include the recommendation of movies (Netflix, Amazon Prime Video), music (Pandora), videos (YouTube), news content (Outbrain) or advertisements (Google). The development of an efficient RS is critical from both the company and the consumer perspective. On one hand, users usually face a very large number of options: for instance, Amazon proposes over 20,000 movies in its selection, and it is therefore important to help them to take the best possible decision by narrowing down the choices they have to make. On the other hand, major companies report significant increase of their traffic and sales coming from personalized recommendations: Amazon declares that $35\%$ of its sales is generated by recommendations, two-thirds of the movies watched on Netflix are recommended and $28\%$ of ChoiceStream users said that they would buy more music, provided the fact that they meet their tastes and interests.\footnote{Talk of Xavier Amatriain - Recommender Systems - Machine Learning Summer School 2014 @ CMU.}

\smallskip


Two main approaches have been proposed to tackle this problem \cite{Ricci:2010:RSH:1941884}. The first
one, referred to as Content-Based recommendation technique \cite{reference/rsh/LopsGS11} makes use of existing contextual information about the users (e.g. demographic information) or items (e.g. textual description) for recommendation. The second approach, referred to as collaborative filtering (CF) and undoubtedly the most popular one, relies on the past interactions and recommends items to users based on the feedback provided by other similar users. Feedback can be {\it explicit}, in the form of ratings; or {\it implicit}, which includes clicks, browsing over an item or listening to a song. Such implicit feedback is readily available in abundance but is more challenging to take into account as it does not clearly depict the preference of a user for an item. Explicit feedback, on the other hand, is very hard to get in abundance.

\smallskip

The adaptation of CF systems designed for another type of feedback has been shown to be sub-optimal as the basic hypothesis of these systems inherently depends on the nature of the feedback \cite{ir2004010}. Further, learning a suitable representation of users and items has been shown to be the bottleneck of these systems \cite{DBLP:conf/kdd/WangWY15}, mostly in the cases where contextual information over users and items which allow to have a richer representation is unavailable.

\bigskip

In this paper we are interested in the learning of user preferences mostly provided in the form of implicit feedback in RS. Our aim is twofold and concerns:
\begin{enumerate}
\item the development of a theoretical framework for learning user preference in recommender systems and its analysis in the worst case where all users provide a minimum of positive/negative feedback;
\item the design of a new neural-network model based on this framework that learns the preference of users over pairs of items and their representations in an embedded space simultaneously without requiring any contextual information.
\end{enumerate}

We extensively validate our proposed approach over standard benchmarks with implicit feedback by comparing it to state of the art models.

The remainder of this paper is organized as follows. In Section \ref{sec:model}, we define the notations and the proposed framework, and analyze its theoretical properties. Then, Section \ref{sec:sim} provides an overview of existing related methods.
Section \ref{sec:experiment} is devoted to numerical experiments on four real-world benchmark data sets including binarized versions of MovieLens and Netflix, and one real data set on online advertising. We compare different versions of our model with state-of-the-art methods showing the appropriateness of our contribution. Finally, we summarize the study and give possible future research perspectives in Section \ref{sec:conclusion}.

\section{User preference and embedding learning with Neural nets }\label{sec:model}

We denote by $\userS\subseteq \N$ (resp. $\itemS\subseteq \N$) the set of indexes over users (resp. the set of indexes over items). 
Further, for each user $u\in\userS$, we consider two subsets of items $\itemS^-_u\subset \itemS$ and $\itemS^+_u\subset \itemS$ such that;
\begin{itemize}
\item[$i)$]  $\itemS^-_u\neq \varnothing$ and $\itemS^+_u \neq \varnothing$,
\item[$ii)$] for any pair of items $(i,i')\in\itemS^+_u\times \itemS^-_u$; $u$ has a preference, symbolized by \!\!$\prefu$\!\!. Hence $i\!\prefu\! i'$ implies that, user $u$ prefers item $i$ over item $i'$.
\end{itemize}
From this preference relation, a desired output $y_{i,u,i'}\in\{-1,+1\}$ is defined over each triplet $(i,u,i')\in\itemS^+_u\times\userS\times\itemS^-_u$ as:

\begin{equation}
y_{i,u,i'}= \left\{
    \begin{array}{ll}
        1 & \mbox{if } i\!\prefu\! i', \\
        -1 & \mbox{otherwise.}
    \end{array}
\right.
\label{eq:Preference}
\end{equation}

\subsection{Learning objective}

The learning task we address is to find a scoring function $f$ from the class of functions $\mathcal F=\{f\mid f: \itemS\times\userS\times \itemS\rightarrow \R\}$ that minimizes the ranking loss:
\begin{equation}
\label{eq:PrefObj}
\Loss(f)=\EE\left[\frac{1}{|\itemS^+_u||\itemS^-_u|}\sum_{i\in\itemS^+_u}\sum_{i'\in\itemS^-_u}\Ind_{y_{i,u,i'}f(i,u,i')<0}\right],
\end{equation}
where $|.|$ measures the cardinality of sets and $\Ind_{\pi}$ is the indicator function which is equal to $1$, if the predicate $\pi$ is true, and $0$ otherwise. Here we suppose that there exists a mapping function $\Phi:\userS\times\itemS\rightarrow \Input\subseteq \mathbb{R}^k$ that projects a pair of user and item indices into a feature space of dimension $k$, and a function $g:\mathcal X\times \mathcal X\rightarrow \R$ such that each function $f\in\mathcal F$ can be decomposed as:
\begin{equation}
\label{eq:deff}
\forall u\in\userS, (i,i')\in\itemS^+_u\times \itemS^-_u,~ f(i,u,i')=g(\Phi(u,i))-g(\Phi(u,i')).
\end{equation}
In the next section we will present a Neural-Network model that learns the mapping function $\Phi$ and outputs the function $f$ based on a non-linear transformation of the user-item feature representation, defining the function $g$.

The previous loss \eqref{eq:PrefObj} is a pairwise ranking loss and it is related to the Area under the ROC curve  \cite{Usunier:1121}. The learning objective is, hence, to find a function $f$ from the class of functions $\mathcal F$ with a small expected risk, by minimizing the empirical error over a training set
\[
S=\{(\bfZ_{i,u,i'}\doteq(i,u,i'),y_{i,u,i'})\mid u\in\userS, (i,i')\in\itemS^+_u\times \itemS^-_u\},
\]
constituted over $N$ users, $\userS=\{1,\ldots,N\}$, and their respective preferences over $M$ items, $\itemS=\{1,\ldots,M\}$ and is given by:
\begin{align}
&\hat\Loss(f,S)=\frac{1}{N}\sum_{u\in\userS}\frac{1}{|\itemS^+_u||\itemS^-_u|}\sum_{i\in\itemS^+_u}\sum_{i'\in\itemS^-_u} \Ind_{y_{i,u,i'}\left(f(i,u,i')\right)<0} \nonumber\\
&=\frac{1}{N}\sum_{u\in\userS}\frac{1}{|\itemS^+_u||\itemS^-_u|}\sum_{i\in\itemS^+_u}\sum_{i'\in\itemS^-_u} \Ind_{y_{i,u,i'}\left(g(\Phi(u,i))-g(\Phi(u,i'))\right)<0}.\label{eq:EmpRisk}
\end{align}
However this minimization problem involves dependent random variables as for each user $u$ and item $i$; all comparisons $g(\Phi(u,i))-g(\Phi(u,i')); i'\in~\itemS^-_u$ involved in the empirical error \eqref{eq:EmpRisk} share the same observation $\Phi(u,i)$. Different studies proposed generalization error bounds for learning with interdependent data \cite{Amini:15}. Among the prominent works that
address this problem are a series of contributions based on the idea of graph coloring introduced in \cite{Janson04RSA}, and which consists in dividing a graph $\graph=(\vertices,\edges)$ that links dependent variables represented by its nodes $\vertices$ into $J$ sets
of {\em independent} variables, called the exact proper fractional cover of $\graph$ and defined as:
\begin{definition}[Exact proper fractional cover of $\graph$, \cite{Janson04RSA}]
\label{def:chromatic}
Let $\graph=(\vertices,\edges)$ be a
graph. $\cover=\{(\Cset_j,\omega_j)\}_{j\in\{1,\ldots,J\}}$, for some positive integer $J$, with
$\Cset_j\subseteq\vertices$ and $\omega_j\in [0,1]$ is an exact proper
fractional cover of $\graph$, if:
i) it is {\em proper:} $\forall j,$ $\Cset_j$ is an {\em independent set}, i.e., there is no connections between vertices in~$\Cset_j$;
ii) it is an {\em exact fractional cover} of $\graph$: $\forall
  v\in\vertices,\;\sum_{j:v\in\Cset_j}\omega_j= 1$.
\end{definition}
The weight $\Weight(\cover)$ of $\cover$ is given by: $\Weight(\cover)\doteq\sum_{j=1}^J\omega_j$ and the
minimum weight $\chi^*(\graph)=\min_{\cover\in\covers(\graph)} \Weight(\cover)$ over the set $\covers(\graph)$ of all exact proper fractional covers of $\graph$ is the {\em fractional chromatic number} of $\graph$.

\begin{figure*}[t!]
\begin{center}
\includegraphics[width=.6\textwidth]{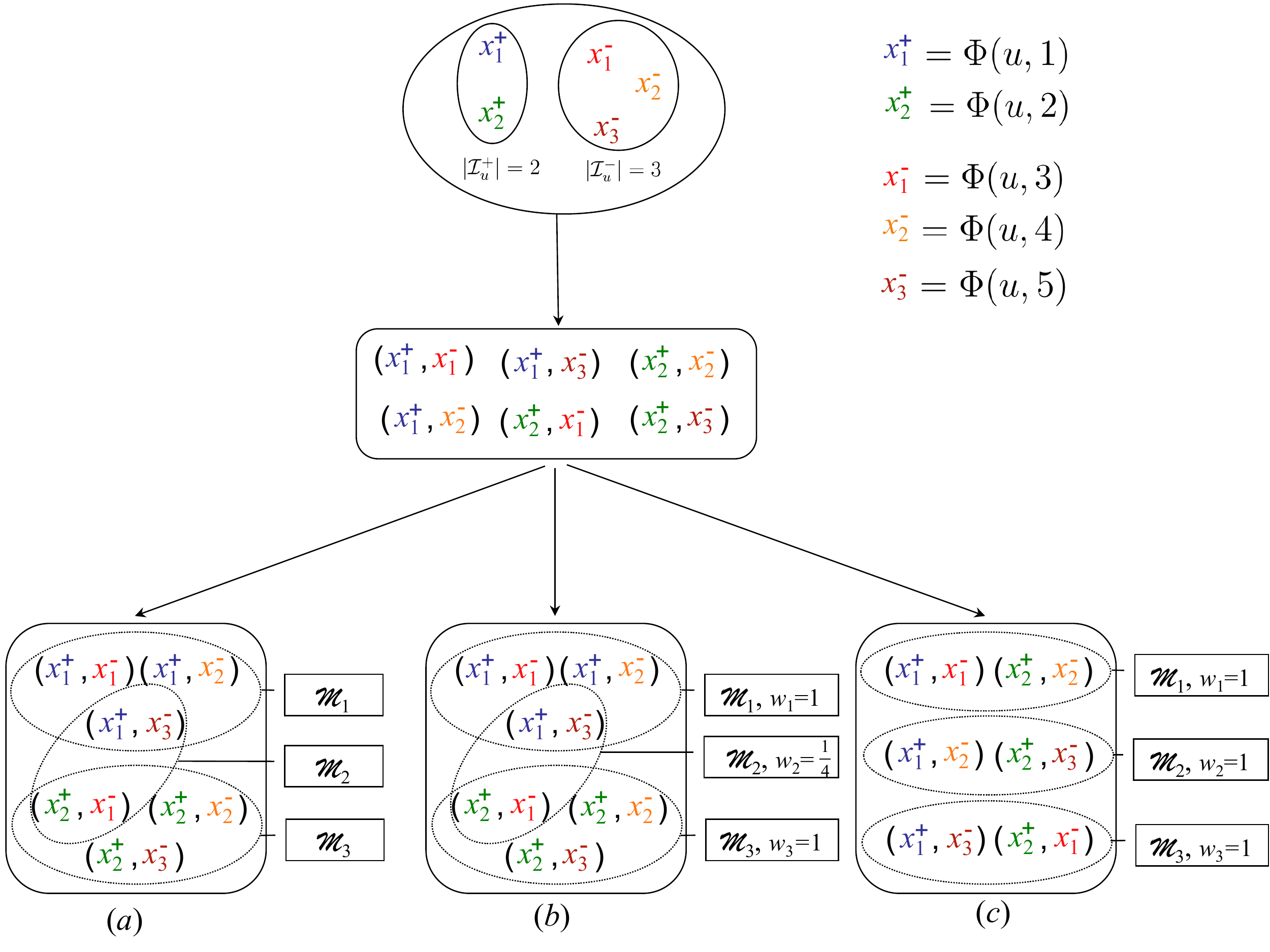}
\end{center}\vspace{-5mm}
\caption{A toy problem with 1 user who prefers $|\itemS_u^+|=2$ items over $|\itemS_u^-|=3$ other ones (top). The dyadic representation of pairs constituted with the representation of the user and each of the representations of preferred and non-preferred items (middle). Different covering of the dependent set, $(a)$ and $(b)$; as well as the exact proper fractional cover, $(c)$, corresponding to the smallest disjoint sets containing independent pairs.}
\label{fig:ProperCover}
\end{figure*}

Figure \ref{fig:ProperCover} depicts an exact proper fractional cover corresponding to the problem we consider for a toy problem with $M=1$ user $u$, and $|\itemS^+_u|=2$ items preferred over $|\itemS^-_u|=3$ other ones. In this case, the nodes of the dependency graph correspond to $6$ pairs constituted by; pairs of the user and each of the preferred items, with the pairs constituted by the user and each of the no preferred items, involved in the empirical loss \eqref{eq:EmpRisk}. Among all the sets containing independent pairs of examples, the one shown in Figure \ref{fig:ProperCover},$(c)$ is the exact proper fractional cover of the $\graph$ and the fractional chromatic number is in this case $\chi^*(\graph)=|\itemS^-_u|=3$.

By mixing the idea of graph coloring with the Laplace transform, Hoeffding like concentration inequalities for the sum of dependent random variables are proposed by~\cite{Janson04RSA}. In \cite{UsunierAG05} this result is extended to provide a generalization of the bounded differences inequality of \cite{mcdiarmid89method} to the case of interdependent random variables. This extension then paved the way for the definition of the {\em fractional Rademacher complexity} that generalizes the idea of Rademacher complexity and allows one to derive generalization bounds for scenarios where the training
data are made of dependent data.

In the worst case scenario where all users provide the lowest interactions over the items, which constitutes the bottleneck of all recommendation systems:
\[
\forall u\in S, |\itemS^-_u|=n_*^-=\mathop{\min}_{u'\in S} |\itemS^-_{u'}| \text{,~and~~} |\itemS^+_u|=n_*^+=\mathop{\min}_{u'\in S} |\itemS^+_{u'}|,
\]

the empirical loss \eqref{eq:EmpRisk} is upper-bounded by:
\begin{equation}
\begin{split}
\label{eq:EmpRisk2}
\hat\Loss(f,S)\le \hat\Loss_*(f,S)\\= \frac{1}{N}\frac{1}{n_*^- n_*^+}\sum_{u\in\userS}\sum_{i\in\itemS^+_u}\sum_{i'\in\itemS^-_u} \Ind_{y_{i,u,i'}f(i,u,i')<0}.
\end{split}
\end{equation}

Following \cite[Proposition 4]{RalaiAmin15}, a generalization error bound can be derived for the second term of the inequality above based on local Rademacher Complexities that implies second-order (i.e. variance) information inducing faster convergence rates.

For sake of presentation and in order to be in line with the learning representations of users and items in an embedded space introduced in Section \ref{sec:RecNetModel}, let us consider kernel-based hypotheses with $\kappa:\Input\times\Input\rightarrow\mathbb{R}$ a {\em positive semi-definite} (PSD) kernel and $\Phi:\userS\times\itemS \rightarrow \Input$ its associated feature mapping function. Further we consider linear functions in the feature space with bounded norm:
\begin{equation}
\mathcal G_B=\{g_{\boldsymbol{w}}\circ \Phi: (u,i)\in \userS\times\itemS \mapsto \langle \boldsymbol{w},\Phi(u,i)\rangle \mid ||\boldsymbol{w}|| \leq B\}
\end{equation}
where $\boldsymbol{w}$ is the weight vector defining the kernel-based hypotheses and $\langle \cdot,\cdot\rangle$ denotes the dot product. We further define the following associated function class:
\begin{equation*}
\resizebox{0.5 \textwidth}{!}{$
\mathcal{F}_B=\{\bfZ_{i,u,i'}\doteq(i,u,i')\mapsto g_{\boldsymbol{w}}(\Phi(u,i))-g_{\boldsymbol{w}}(\Phi(u,i'))\mid g_{\boldsymbol{w}}\in \mathcal G_B\},
$}
\end{equation*}

and the parameterized family $\mathcal{F}_{B,r}$ which, for $r>0$, is defined as:
\[
    \mathcal{F}_{B,r} =
    \{f:f\in\mathcal{F}_B,\mathbb{V}[f]\doteq\mathbb{V}_{\bfZ,y}[\Ind_{y f(\bfZ)}]\leq r\},
\]
where $\mathbb{V}[.]$ denotes the variance.
The fractional Rademacher complexity introduced in \cite{UsunierAG05} entails our analysis:
	    \[
	    \rademacher_{S}(\mathcal{F})=\frac{2}{m}\mathbb{E}_{\xi}\sum_{j=1}^{n_*^-}\mathbb{E}_{\Cset_j}\sup_{f\in\mathcal{F}}\sum_{\alpha\in\Cset_j \atop \bfZ_\alpha \in S}\xi_\alpha f(\bfZ_\alpha),
	\]
	where $m=N\times n_*^+\times n_*^-$ is the total number of triplets $\bfZ$ in the training set and $(\xi_i)_{i=1}^m$ is a sequence of 
	independent Rademacher variables verifying
	$\mathbb{P}(\xi_i=1)=\mathbb{P}(\xi_i=-1)=\frac{1}{2}$.

\bigskip

\begin{theorem}
	\label{thm:WorseCaseRecNet}
        Let $\userS$ be a set of $M$ independent users, such that each user $u \in \userS $ prefers $n_*^+$ items over $n_*^-$ ones in a predefined set of $\itemS$ items. Let $S=\{(\bfZ_{i,u,i'}\doteq(i,u,i'),y_{i,u,i'})\mid u\in\userS, (i,i')\in\itemS^+_u\times \itemS^-_u\}$ be the associated training set, then for any $1>\delta>0$ the following generalization bound holds for all $f\in  \mathcal{F}_{B,r}$ with probability at least $1-\delta$:
        \begin{align*}
     \Loss(f)\le ~~&\hat\Loss_*(f,S) + \frac{2B\mathfrak{C}(S)}{Nn^+_*}+\\ &\frac{5}{2}\left(\sqrt{\frac{2B\mathfrak{C}(S)}{Nn^+_*}}+\sqrt{\frac{r}{2}}\right)\sqrt{\frac{\log\frac{1}{\delta}}{n_*^+}}+\frac{25}{48}\frac{\log\frac{1}{\delta}}{n_*^+},
     \end{align*}
    where $\mathfrak{C}(S)=\sqrt{ \frac{1}{n^-_*}\sum_{j=1}^{n_*^-}\mathbb{E}_{\Cset_j}\left[  \sum_{\alpha\in\Cset_j \atop \bfZ_\alpha \in S}d(\bfZ_\alpha,\bfZ_{\alpha}))\right]}$, $\bfZ_\alpha=(i_\alpha,u_\alpha,i'_\alpha)$ and \\ \begin{align*}
    d(\bfZ_\alpha,\bfZ_{\alpha})=~&\kappa(\Phi(u_\alpha,i_\alpha),\Phi(u_\alpha,i_\alpha))\\+\kappa(\Phi(u_\alpha,i'_\alpha),\Phi(u_\alpha,i'_\alpha))-
    &2\kappa(\Phi(u_\alpha,i_\alpha),\Phi(u_\alpha,i'_\alpha)).
    \end{align*}
\end{theorem}

The proof is given in Appendix. This result suggests that~:
\begin{itemize}
\item even though the training set $S$ contains interdependent observations; following \cite[theorem 2.1, p. 38]{vapnik2000nature}, theorem \ref{thm:WorseCaseRecNet} gives insights on the consistency of the empirical risk minimization principle with respect to \eqref{eq:EmpRisk2},
\item in the case where the feature space $\Input\subseteq \mathbb{R}^k$ is of finite dimension; lower values of $k$ involves lower kernel estimation and hence lower complexity term $\mathfrak{C}(S)$ which implies a tighter generalization bound.
\end{itemize}

\subsection{A Neural Network model to learn user preference}
\label{sec:RecNetModel}
Some studies proposed to find the dyadic representation of users and items in an embedded space, using neighborhood similarity information \cite{Volkovs:2015} or the Bayesian Personalized Ranking (BPR) \cite{rendle_09}. In this section we propose a feed-forward Neural Network, denoted as {\RecNet}, to learn jointly the embedding representation, $\Phi(.)$, as well as the scoring function, $f(.)$, defined previously. The input of the network is a triplet $(i,u,i')$ composed by the indexes of an item $i$, a user $u$ and a second item $i'$; such that the user $u$ has a preference over the pair of items $(i, i')$ expressed by the desired output $y_{i,u,i'}$, defined with respect to the preference relation $\!\prefu\!$ (Eq. \ref{eq:Preference}). Each index in the triplet is then transformed to a corresponding binary indicator vector $\mathbf{i}, \mathbf{u},$ and $\mathbf{i}'$ having all its characteristics equal to $0$ except the one that indicates the position of the user or the items in its respective set, which is equal to $1$. Hence, the following one-hot vector corresponds to the binary vector representation of user $u\in\userS$:
\begin{center}
\includegraphics[width=0.3\textwidth]{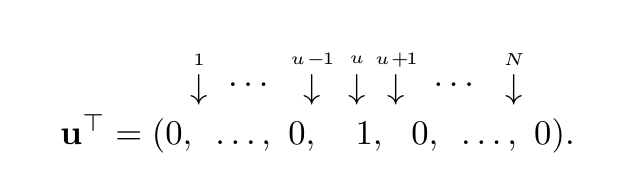}
\end{center}
\bigskip

\begin{figure*}[!ht]
\begin{center}
\includegraphics[width=.7\textwidth]{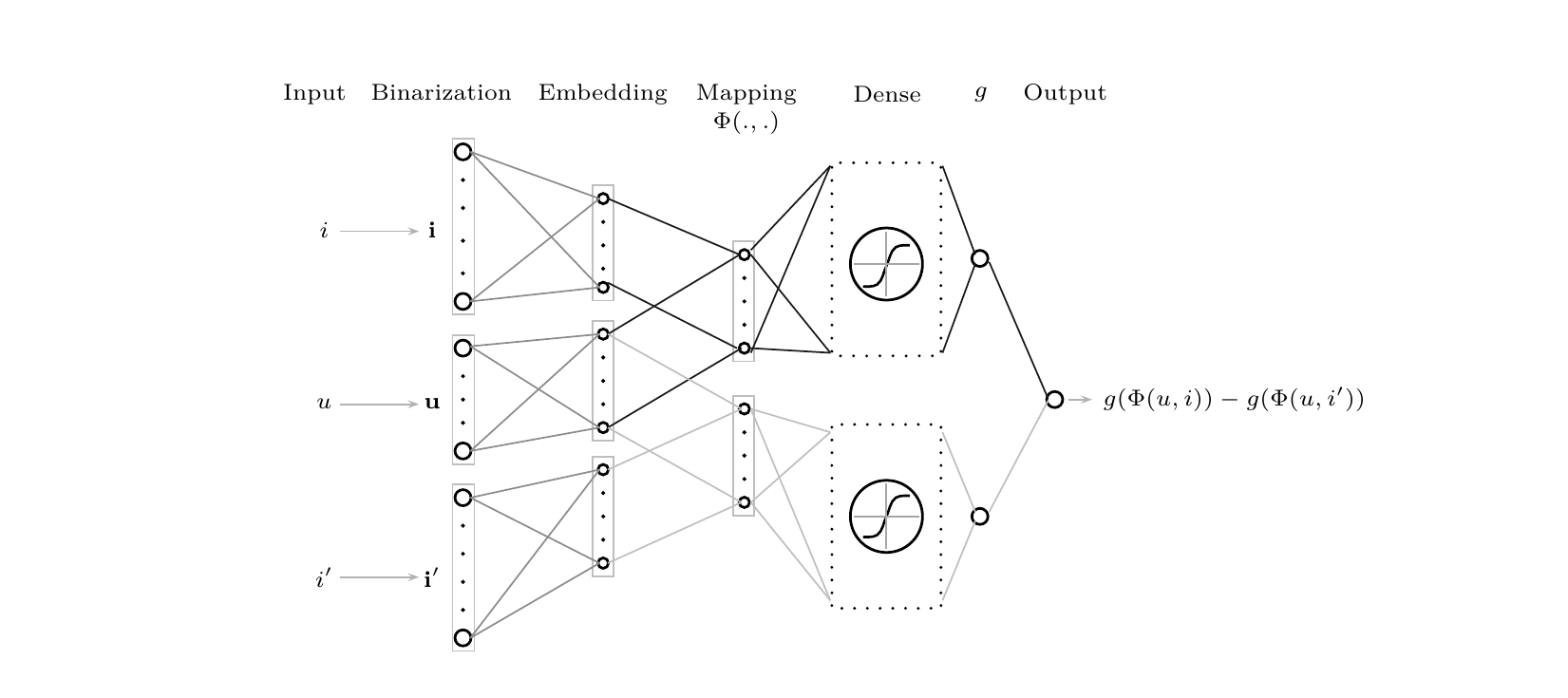}
\end{center}\vspace{-5mm}
\caption{The architecture of {\RecNet} trained to reflect the preference of a user $u$ over a pair of items $i$ and $i'$. }
\label{fig:recnet}
\end{figure*}

The network entails then three successive layers, namely {\it Embedding} (SG), {\it Mapping} and {\it Dense} hidden layers depicted in Figure  \ref{fig:recnet}.

\begin{itemize}
   \item The {\it Embedding} layer transforms the sparse binary representations of the user and each of the items to a denser real-valued vectors. We denote by $\vecU_u$ and $\vecI_i$ the transformed vectors of user $u$ and item $i$; and $\mathbf{\mathbb{U}}=(\vecU_u)_{u\in\userS}$ and $\mathbf{\mathbb{V}}=(\vecI_i)_{i\in\itemS}$ the corresponding matrices. Note that as the binary indicator vectors of users and items contain one single non-null characteristic, each entry of the corresponding dense vector in the SG layer is connected by only one weight to that characteristic.
   \item The {\it Mapping} layer is composed of two groups of units each being obtained from the element-wise product between the user representation vector $\vecU_u$ of a user $u$ and a corresponding item representation vector $\vecI_i$ of an item $i$ inducing the feature representation of the pair $(u,i); \Phi(u,i)$.  
    \item Each of these units are also fully connected to the units of a {\it Dense} layer composed of successive hidden layers (see Section \ref{sec:experiment} for more details related to the number of hidden units and the activation function used in this layer) .
\end{itemize}

The model is trained such that the output of each of the dense layers reflects the relationship between the corresponding item and the user and is mathematically defined by a multivariate real-valued function $g(.)$.
Hence, for an input $(i,u,i')$, the output of each of the dense layers is a real-value score that reflects a preference associated to the corresponding pair $(u,i)$ or $(u,i')$ (i.e. $g(\Phi(u,i))$ or $g(\Phi(u,i'))$). Finally the prediction given by {\RecNet} for an input $(i,u,i')$ is:
\begin{equation}
\label{eq:defF}
f(i,u,i')=g(\Phi(u,i))-g(\Phi(u,i')).
\end{equation}


\subsection{Algorithmic implementation}

We decompose the ranking loss as a linear combination of two logistic surrogates:
   \begin{equation}
   \label{eq:rankingLoss}
   \mathnormal
   \Loss_{c,p}(f,\mathbb{U},\mathbb{V},\Trn)=\Loss_c(f,\Trn)+\Loss_p(\mathbb{U},\mathbb{V},\Trn),
    \end{equation}
   where the first term reflects the ability of the non-linear transformation of user and item feature representations, $g(\Phi(.,.))$, to respect the relative ordering of items with respect to users' preferences:
     \begin{equation}
     \resizebox{0.5\textwidth}{!}{$
    \label{eq:ranking_loss}
    \Loss_c(f,\Trn)=\frac{1}{|\Trn|}\sum_{(\bfZ_{i,u,i'},y_{i,u,i'})\in \Trn} \log(1+e^{y_{i,u,i'}(g(\Phi(u,i'))-g(\Phi(u,i))}).
    $}
    \end{equation}
    The second term focuses on the quality of the compact dense vector representations of items and users that have to be found, as measured by the ability of the dot-product in the resulting embedded vector space to respect the relative ordering of preferred items by users:

    \begin{equation}
    \resizebox{0.5\textwidth}{!}{$
    \label{eq:embedding_loss}
    \Loss_p(\mathbb{U},\mathbb{V},\Trn)=\frac{1}{|\Trn|}\!\!\sum_{(\bfZ_{i,u,i'},y_{i,u,i'})\in\Trn}\!\!\left[\log(1+e^{y_{i,u,i'}\vecU_u^\top(\vecI_{i'}-\vecI_{i})})+\lambda(\|\vecU_u\|+\|\vecI_{i'}\|+\|\vecI_{i}\|)\right],
    $}
    \end{equation}

    where $\lambda$ is a regularization parameter for the user and items norms.
Finally, one can also consider a version in which both losses are assigned different weights:
    \begin{equation}\label{eq:rankingLoss_alpha}
   \mathnormal
   \Loss_{c,p}(f,\mathbb{U},\mathbb{V},\Trn)=\alpha\Loss_c(f,\Trn)+(1-\alpha)\Loss_p(\mathbb{U},\mathbb{V},\Trn),
   \end{equation}
where $\alpha\in [0,1]$ is a real-valued parameter to balance between ranking prediction ability and expressiveness of the learned item and user representations. Both options will be discussed in the experimental section.

\subsubsection*{Training phase}

The training of the {\RecNet} is done by back-propagating \cite{Leon2012} the error-gradients from the output to both the deep and embedding parts of the model using mini-batch stochastic optimization (Algorithm 1).


During training, the input layer takes a random set $\tilde \Trn_n$ of size $n$ of interactions by building triplets $(i,u,i')$ based on this set, and generating a sparse representation from id's vector corresponding to the picked user and the pair of items. The binary vectors of the examples in  $\tilde \Trn_n$ are then propagated throughout the network, and the ranking error (Eq. \ref{eq:rankingLoss}) is back-propagated.

\begin{algorithm}[!ht]
\caption{{\RecNet$_.$}: Learning phase}
\begin{algorithmic}[J]
\REQUIRE \STATE $T$: maximal number of epochs
\STATE A set of users $\userS=\{1,\ldots,N\}$
\STATE A set of items $\itemS=\{1,\ldots,M\}$

\FOR{$ep=1,\dots,T$}
\STATE Randomly sample a mini-batch $\tilde \Trn_n\subseteq \Trn$ of size $n$ from the original user-item matrix
\FORALL{$((i,u,i'),y_{i,u,i'})\in \tilde \Trn_n$}
    	\STATE \textbf{Propagate} $(i,u,i')$ from the input to the output.
\ENDFOR
    	\STATE \textbf{Retro-propagate} the pairwise ranking error (Eq. \ref{eq:rankingLoss}) estimated over $\tilde \Trn_n$.
\ENDFOR
\ENSURE Users and items latent feature matrices $\mathbb{U}, \mathbb{V}$ and the model weights.
\end{algorithmic}
\end{algorithm}

\subsubsection*{Model Testing}
As for the prediction phase, shown in Algorithm 2, a ranked list $\mathfrak N_{u,k}$ of the $k\ll M$ preferred items for each user in the test set is maintained while retrieving the set $\mathcal I$. Given the latent representations of the triplets, and the weights learned; the two first items in $\mathcal I$ are placed in $\mathfrak N_{u,k}$ in a way which ensures that preferred one, $i^*$, is in the first position. Then, the algorithm retrieves the next item, $i\in \mathcal I$ by comparing it to $i^*$. This step is simply carried out by comparing the model's output over the concatenated binary indicator vectors of $(i^*, u, i)$ and $(i, u, i^*)$.

\medskip

\begin{algorithm}[b!]
\caption{{\RecNet$_.$}: Testing phase}
\begin{algorithmic}[J]
\REQUIRE \STATE A user $u\in\userS$; A set of items $\itemS=\{1,\ldots,M\}$; \\
A set containing the $k$ preferred items in $\itemS$ by $u$;\\
$\mathfrak N_{u,k} \leftarrow \varnothing$;
\STATE The output of {\RecNet}$_{.}$ learned over a training set: $f$
\STATE Apply $f$ to the first two items of $\mathcal I$ and, note the preferred one $i^*$ and place it at the top of $\mathfrak N_{u,k}$;
\FOR{$i=3,\dots,M$}
		\IF {$g(\Phi(u,i))>g(\Phi(u,i^*))$}
			\STATE Add $i$ to $\mathfrak N_{u,k}$ at rank 1
		\ELSE
		    \STATE $j\leftarrow 1$
			\WHILE {$j\le k$ AND $g(\Phi(u,i))<g(\Phi(u,i_g))$) {\color{gray} // where $i_g=\mathfrak N_{u,k}(j)$}}
			    \STATE $j\leftarrow j+1$
			\ENDWHILE
            \IF {$j\le k$}
              \STATE Insert $i$ in $\mathfrak N_{u,k}$ at rank $j$
            \ENDIF
        \ENDIF
\ENDFOR
\ENSURE $\mathfrak N_{u,k}$;
\end{algorithmic}
\end{algorithm}

Hence, if $f(i,u,i^*)>f(i^*,u,i)$, which from Equation \ref{eq:defF} is equivalent to $g(\Phi(u,i))~>~g(\Phi(u,i^*))$, then $i$ is predicted to be preferred over $i^*$; $i \prefu i^*$; and it is put at the first place instead of $i^*$ in $\mathfrak N_{u,k}$. Here we assume that the predicted preference relation \!\!$\prefu$\!\! is transitive, which then ensures that the predicted order in the list is respected. Otherwise, if $i^*$ is predicted to be preferred over $i$, then $i$ is compared to the second preferred item in the list, using the model' prediction as before, and so on. The new item, $i$, is inserted in $\mathfrak N_{u,k}$ in the case if it is found to be preferred over another item in $\mathfrak N_{u,k}$.

\medskip

By repeating the process until the end of $\mathcal I$, we obtain a ranked list of the $k$ most preferred items for the user $u$. Algorithm 2 does not require an ordering of the whole set of items, as also in most cases we are just interested in the relevancy of the top ranked items for assessing the quality of a model. Further, its complexity is at most $O(k\times M)$ which is convenient in the case where $M>\!\!>\! 1$. The merits of a similar algorithm have been discussed by \cite{Ailon08anefficient} but, as pointed out above, the basic assumption for inserting a new item in the ranked list $\mathfrak N_{u,k}$ is that the predicted preference relation induced by the model should be transitive, which may not hold in general.

\smallskip

In our experiments, we also tested a more conventional inference algorithm, which for a given user $u$, consists in the ordering of items in $\mathcal I$ with respect to the output given by the function $g$, and we did not find any substantial difference in the performance of {\RecNet}$_.$, as presented in the following section.

\section{(Un)-related work}\label{sec:sim}

This section provides an overview of the state-of-the-art approaches that are the most similar to ours.

\subsection{Neural Language Models}
Neural language models have proven themselves to be successful in many natural language processing tasks including speech recognition, information retrieval and sentiment analysis. These models are based on a distributional hypothesis stating that words, occurring in the same context with the same frequency, are similar. In order to capture such similarities, these approaches propose to embed the word distribution into a low-dimensional continuous space using Neural Networks, leading to the development of several powerful and highly scalable language models such as the word2Vec Skip-Gram (SG) model \cite{word_emb,mikolov_13}.


\medskip

The recent work of \cite{levy_14} has shown new opportunities to extend the word representation learning to characterize more complicated pieces of information. In fact, this paper established the equivalence between SG model with negative sampling, and implicitly factorizing a point-wise mutual information (PMI) matrix. Further, they demonstrated that word embedding can be applied to different types of data, provided that it is possible to design an appropriate context matrix for them. This idea has been successfully applied to recommendation systems where different approaches attempted to learn representations of items and users in an embedded space in order to meet the problem of recommendation more efficiently \cite{guardia_15, liang_16, DBLP:conf/kdd/GrbovicRDBSBS15}.


\medskip

In \cite{ He:2017:NCF:3038912.3052569}, the authors
used a bag-of-word vector representation of items and users, from which the latent representations of latter are learned through word-2-vec.
\cite{liang_16} proposed a model that relies on the intuitive idea that the pairs of items which are scored in the same way by different users are similar. The approach reduces to finding both the latent representations of users and items, with the traditional Matrix Factorization (MF) approach, and simultaneously learning item embeddings using a co-occurrence shifted positive PMI (SPPMI) matrix defined by items and their context. The latter is used as a regularization term in the traditional objective function of MF. Similarly, in \cite{DBLP:conf/kdd/GrbovicRDBSBS15} the authors proposed Prod2Vec, which embeds items using a Neural-Network language model applied to a time series of user purchases. This model was further extended in \cite{vasile_16} who, by defining appropriate context matrices, proposed a new model called Meta-Prod2Vec. Their approach learns a representation for both items and side information available in the system. The embedding of additional information is further used to regularize the item embedding.
Inspired by the concept of sequence of words; the approach proposed by \cite{guardia_15} defined the consumption of items by users as trajectories. Then, the embedding of items is learned using the SG model and the users' embeddings are further used to predict the next item in the trajectory. In these approaches, the learning of item and user representations are employed to make prediction with predefined or fixed similarity functions (such as dot-products) in the embedded space.

\subsection{Learning-to-Rank with Neural Networks}
Motivated by automatically tuning the parameters involved in the combination of different scoring functions, Learning-to-Rank approaches were originally developed for Information Retrieval (IR) tasks and are grouped into three main categories: pointwise, listwise and pairwise \cite{Liu:2009}.

\medskip

Pointwise approaches \cite{Crammer01, Li08} assume that each queried document pair has an ordinal score. Ranking is then formulated as a regression problem, in which the rank value of each document is estimated as an absolute quantity. In the case
where relevance judgments are given as pairwise preferences
(rather than relevance degrees), it is usually not straightforward to apply these algorithms for learning. Moreover, pointwise techniques do not consider the inter-dependency among documents, so that the position of documents in the final ranked list is missing in the regression-like loss functions used for parameter tuning. On the other hand, listwise approaches \cite{Shi:2010, Xu07, Xu08} take the entire ranked
list of documents for each query as a training instance. As
a direct consequence, these approaches are able to differentiate documents from different queries, and consider their position in the output ranked list at the training stage. Listwise techniques aim to directly optimize a ranking
measure, so they generally face a complex optimization problem dealing with non-convex, non-differentiable and discontinuous functions. Finally, in pairwise approaches \cite{Cohen99, Freund03, Joachims02, PessiotTUAG07} the ranked list is decomposed into a set of document pairs. Ranking is therefore considered as the classification of pairs of documents, such that a classifier is trained by minimizing the number of misorderings in ranking. In the test phase, the classifier assigns a positive or negative class label to a document pair that indicates which of the documents in the pair should be better ranked than the other one.

\smallskip

Perhaps the first Neural Network model for ranking is RankProp, originally proposed by \cite{Caruana:1995}. RankProp is a pointwise approach that alternates between two phases of learning the desired real outputs by minimizing a Mean Squared Error (MSE) objective, and a modification of the desired values themselves to reflect the current ranking given by the net. Later on \cite{DBLP:conf/icml/BurgesSRLDHH05} proposed RankNet, a pairwise approach, that learns a preference function by minimizing a cross entropy cost over the pairs of relevant and irrelevant examples. SortNet proposed by \cite{DBLP:conf/icann/RigutiniPMB08, DBLP:journals/tnn/RigutiniPMS11} also learns a preference function by minimizing a ranking loss over the pairs of examples that are selected iteratively with the overall aim of maximizing the quality of the ranking. The three approaches above consider the problem of Learning-to-Rank for IR and without learning an embedding.


\section{Experimental Results}\label{sec:experiment}

We conducted a number of experiments aimed at evaluating how the simultaneous learning of user and item representations, as well as the preferences of users over items can be efficiently handled with {\RecNet}$_.$. To this end, we considered four real-world benchmarks commonly used for collaborative filtering. We validated our approach with respect to different hyper-parameters that impact the accuracy of the model and compare it with competitive state-of-the-art approaches.

\medskip

We run all experiments on a cluster of five {32 core Intel Xeon @ 2.6Ghz CPU (with 20MB cache per core)} systems with {256 Giga} RAM running {Debian GNU/Linux 8.6 (wheezy)} operating system.\cmmnt{Finally, since \NetF\ and \kasandr\ data sets are quite large, we run experiments on these data sets on 2 GRID-GPU(s) each having 8 GPU(s) of their own with {4 Giga} RAM.}
All subsequently discussed components were implemented in Python3 using the TensorFlow library with version 1.4.0.\footnote{\url{https://www.tensorflow.org/}. }$^,$\footnote{For research purpose we will make available all the codes implementing Algorithms 1 and 2 that we used in our experiments and all the pre-processed datasets.}
\begin{figure*}[!t]
    \centering
       \subfloat[{\ML-100K}]{
    \includegraphics[width=0.4\textwidth]{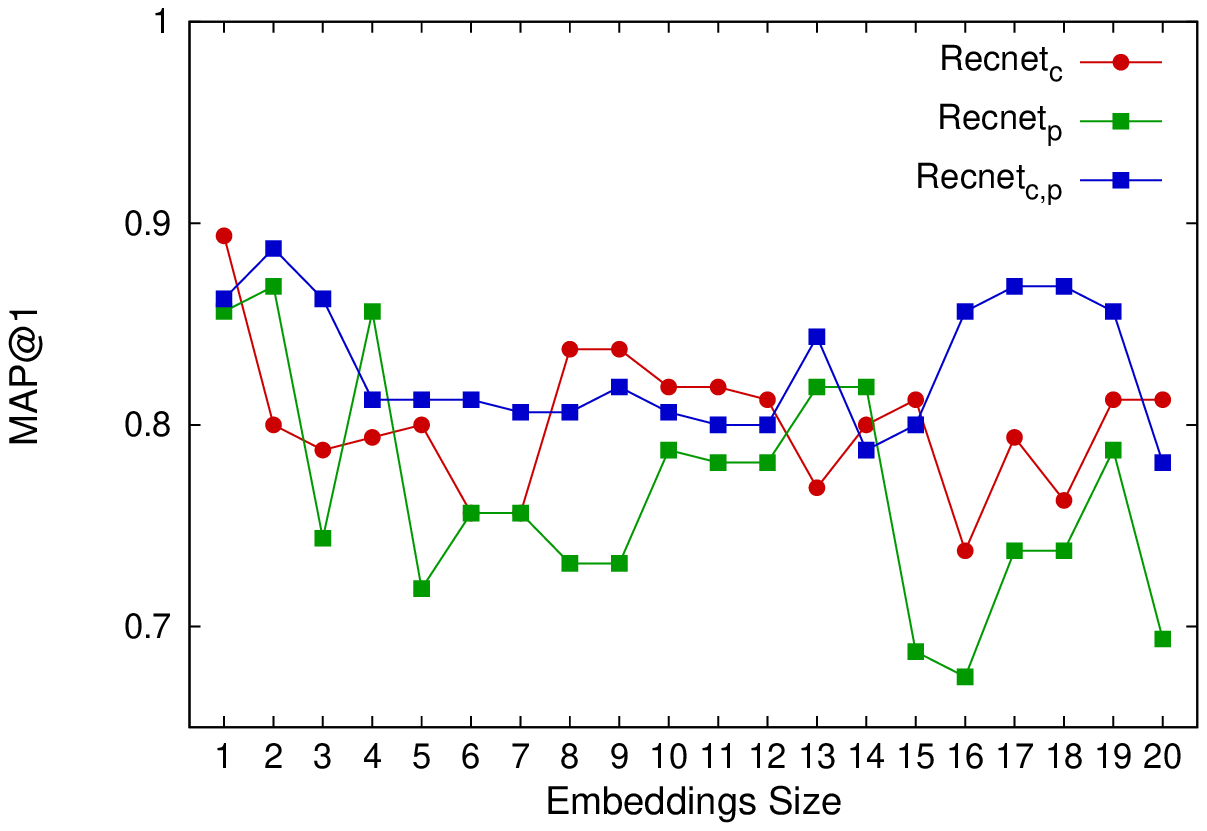}
    }
    \subfloat[{\ML-1M}]{
    \includegraphics[width=0.4\textwidth]{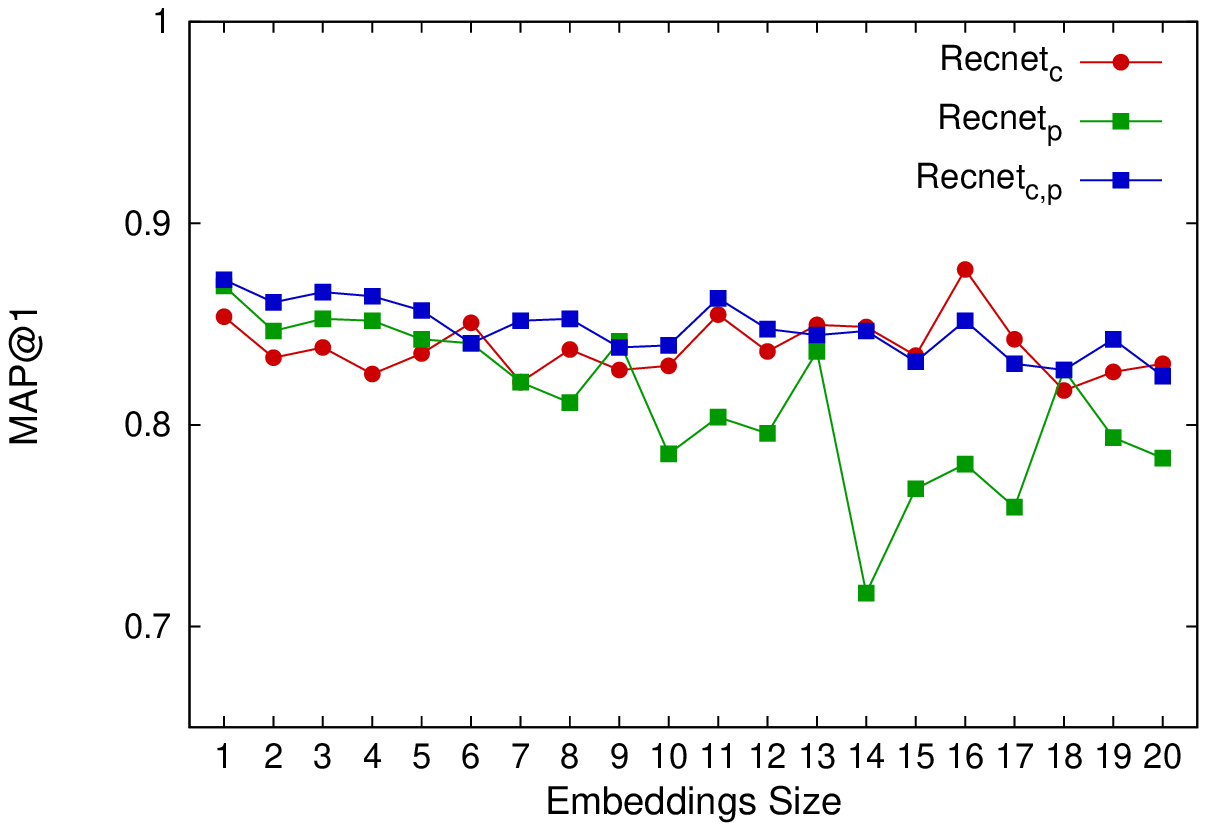}
    }\vspace{-2mm}\\
        \subfloat[{\kasandr}]{
    \includegraphics[width=0.4\textwidth]{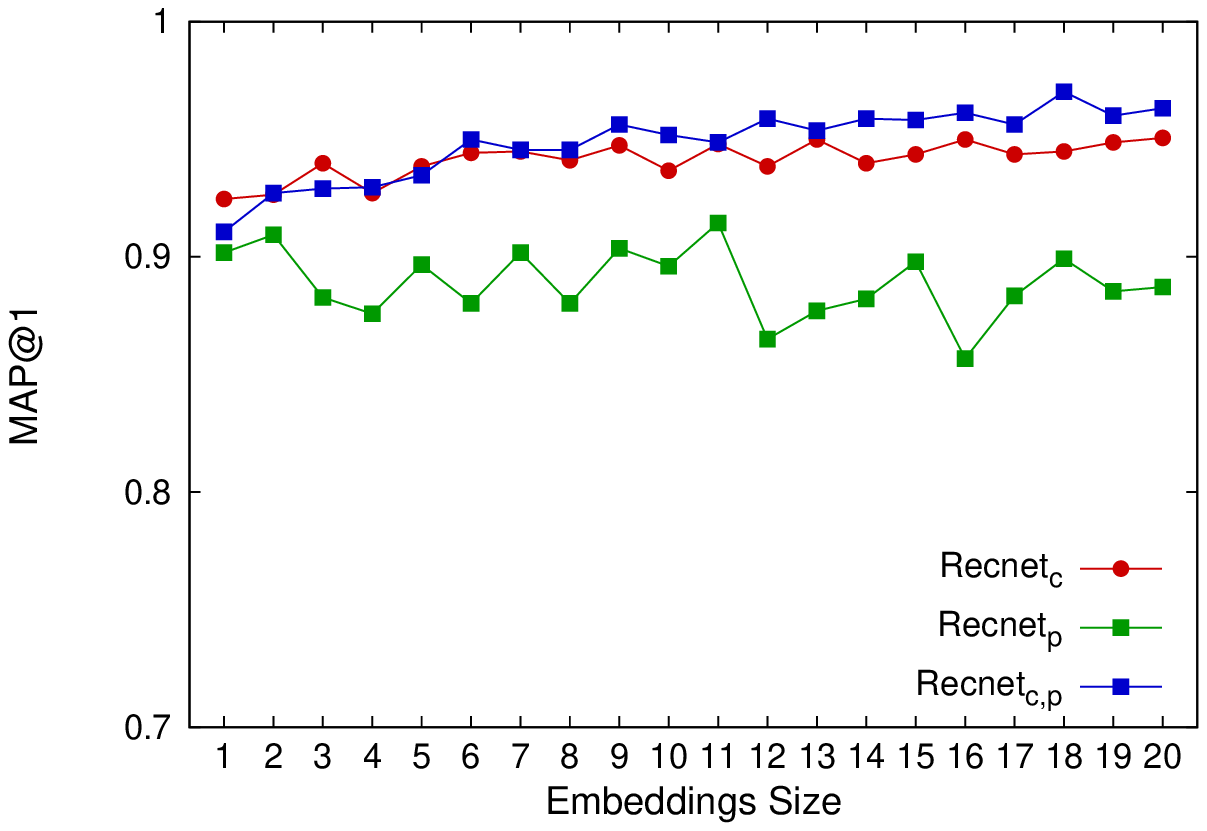}
    }
            \subfloat[{\NetF{}}]{
    \includegraphics[width=0.4\textwidth]{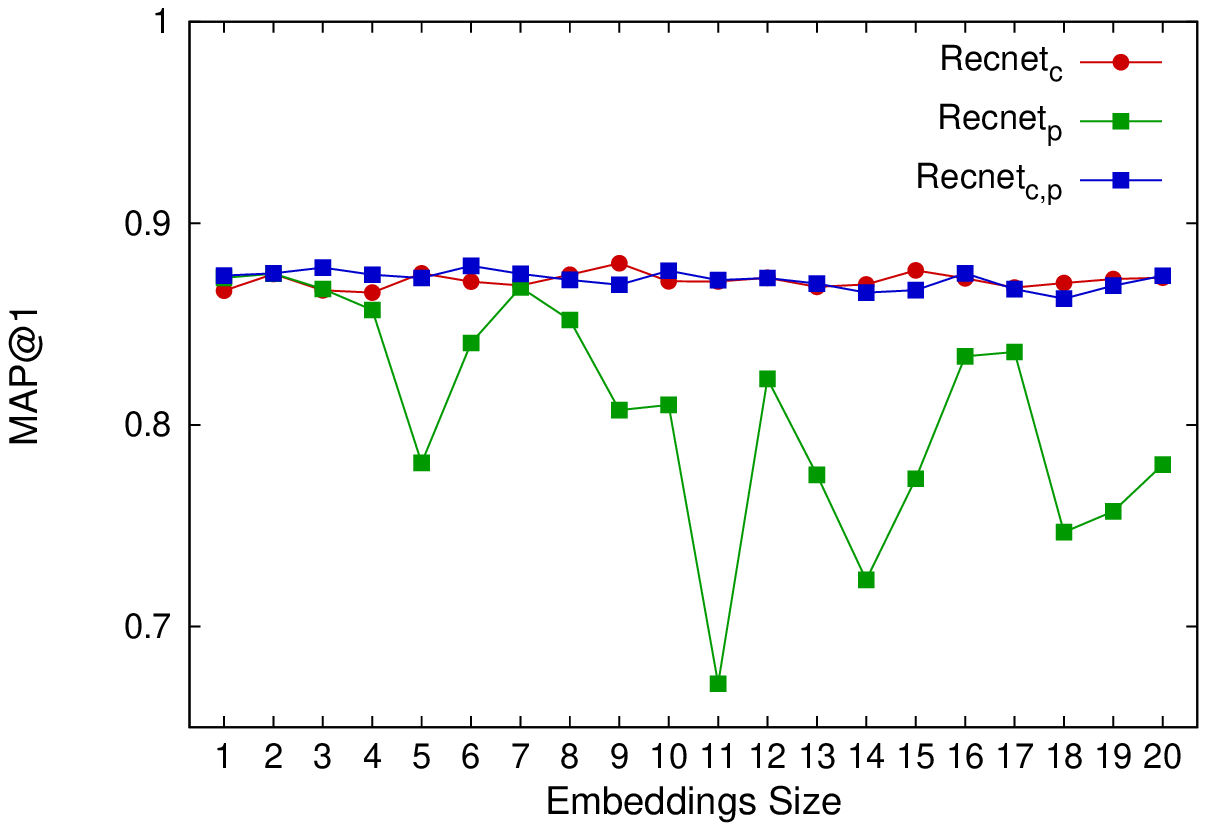}
    }\vspace{-1mm}
    \caption{MAP@1 as a function of the dimension of the embedding for {\ML}-100K, {\ML}-1M and {\kasandr}.}
    \label{fig:emb_dimension}
\end{figure*}

\subsection{Datasets}
\label{sec:Data}
We report results obtained on three publicly available movie data\-sets, for the
task of personalized top-N recommendation: {\MovieL}\footnote{\url{https://movielens.org/}} 100K (\ML-100K), {\MovieL} 1M (\ML-1M) \cite{Harper:2015:MDH:2866565.2827872}, {\NetF}\footnote{\url{http://academictorrents.com/details/9b13183dc4d60676b773c9e2cd6de5e5542cee9a}}, and one clicks dataset, {\kasandr}-Germany \footnote{\url{https://archive.ics.uci.edu/ml/datasets/KASANDR}} \cite{DBLP:conf/sigir/sidana17}, a recently released data set for on-line advertising.
\begin{itemize}
\item \ML-100K, \ML-1M and {\NetF} consists of user-movie ratings, on a scale of one to five, collected from a movie recommendation service and the Netflix company. The latter was released to support the Netlfix Prize competition\footnote{B. James and L. Stan, The Netflix Prize (2007).}. \cmmnt{\ML-100K dataset gathers 100,000 ratings from 943 users on 1682 movies, \ML-1M dataset comprises of 1,000,000 ratings from 6040 users and 3900 movies and {\NetF} consists of 100 million ratings from 480,000 users and 17,000 movies.} For all three datasets, we only keep users who have rated at least five movies and remove users who gave the same rating for all movies. In addition, for {\NetF}, we take a subset of the original data and randomly sample $20\%$ of the users and $20\%$ of the items. In the following experiments, as we only compare with approaches developed for the ranking purposes and our model is designed to handle implicit feedback, these three data sets are made binary such that a rating higher or equal to 4 is set to 1 and to 0 otherwise.


\vspace{1mm}\item The original {\kasandr} dataset contains the interactions and clicks done by the users of Kelkoo, an online advertising platform, across twenty Europeans countries. In this article, we used a subset of {\kasandr} that only considers interactions from Germany. It gathers 17,764,280 interactions from 521,685 users on 2,299,713 offers belonging to 272 categories and spanning across 801 merchants. For \kasandr, we remove users who gave the same rating for all offers. This implies that all the users who never clicked\cmmnt{ (and always had negative rating on all offers) } or always clicked on each and every offer shown to them\cmmnt{ (and always had positive rating on all offers)} were removed.
\end{itemize}

Table \ref{tab:dataset-description} provides the basic statistics on these collections after pre-processing, as discussed above.

\begin{table}[!htpb]
\tiny
\centering
\caption{Statistics of various collections used in our experiments after preprocessing.}\vspace{-2mm}
\label{tab:dataset-description}
    \resizebox{0.5\textwidth}{!}{
    \begin{tabular}{lllll}
    \hline
         & \# of users &\# of items & \# of interactions & Sparsity\\
         \hline
        \ML-100K & 943 & 1,682 & 100,000& 93.685\%\\
        \ML-1M   & 6,040 & 3,706 & 1,000,209 & 95.530\% \\
        \NetF & 90,137 & 3,560 & 4,188,098 & 98.700\% \\
        \kasandr&25,848&1,513,038&9,489,273&99.976\%\\
        \hline
    \end{tabular}
    }
\end{table}

\subsection{Experimental set-up}
\begin{figure*}[!htpb]
    \centering
    \subfloat[{\ML-100K}]{
    \includegraphics[width=0.4\textwidth]{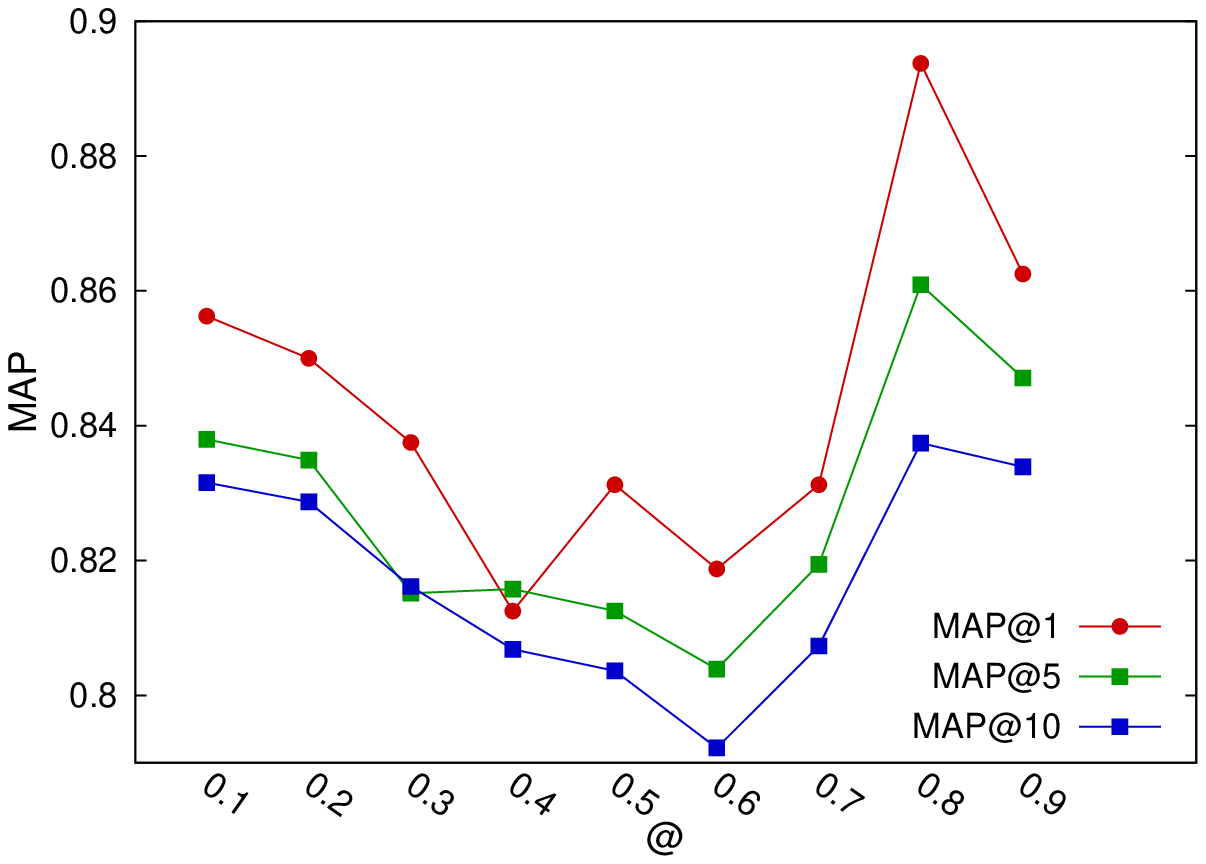}
    }
        \subfloat[{\ML-1M}]{
    \includegraphics[width=0.4\textwidth]{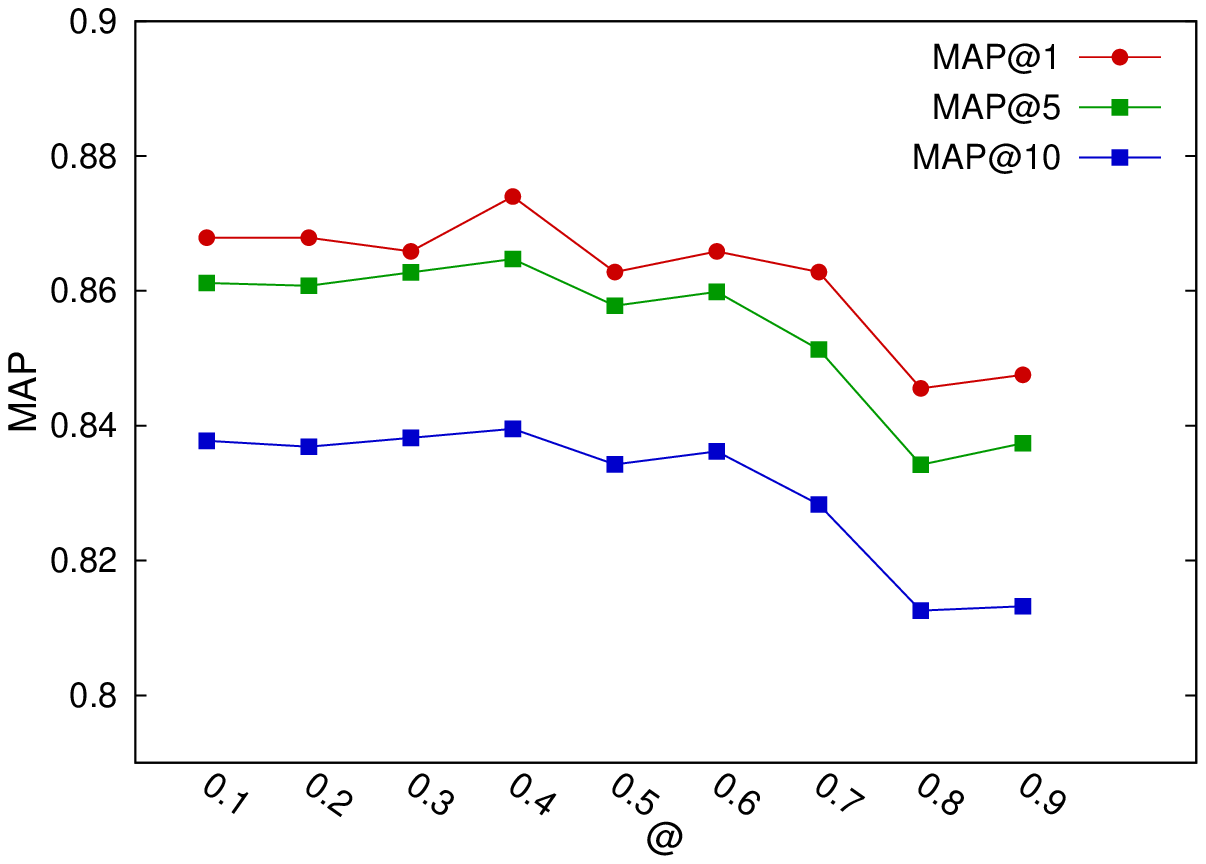}
    }\\
       \subfloat[{\kasandr}]{
    \includegraphics[width=0.4\textwidth]{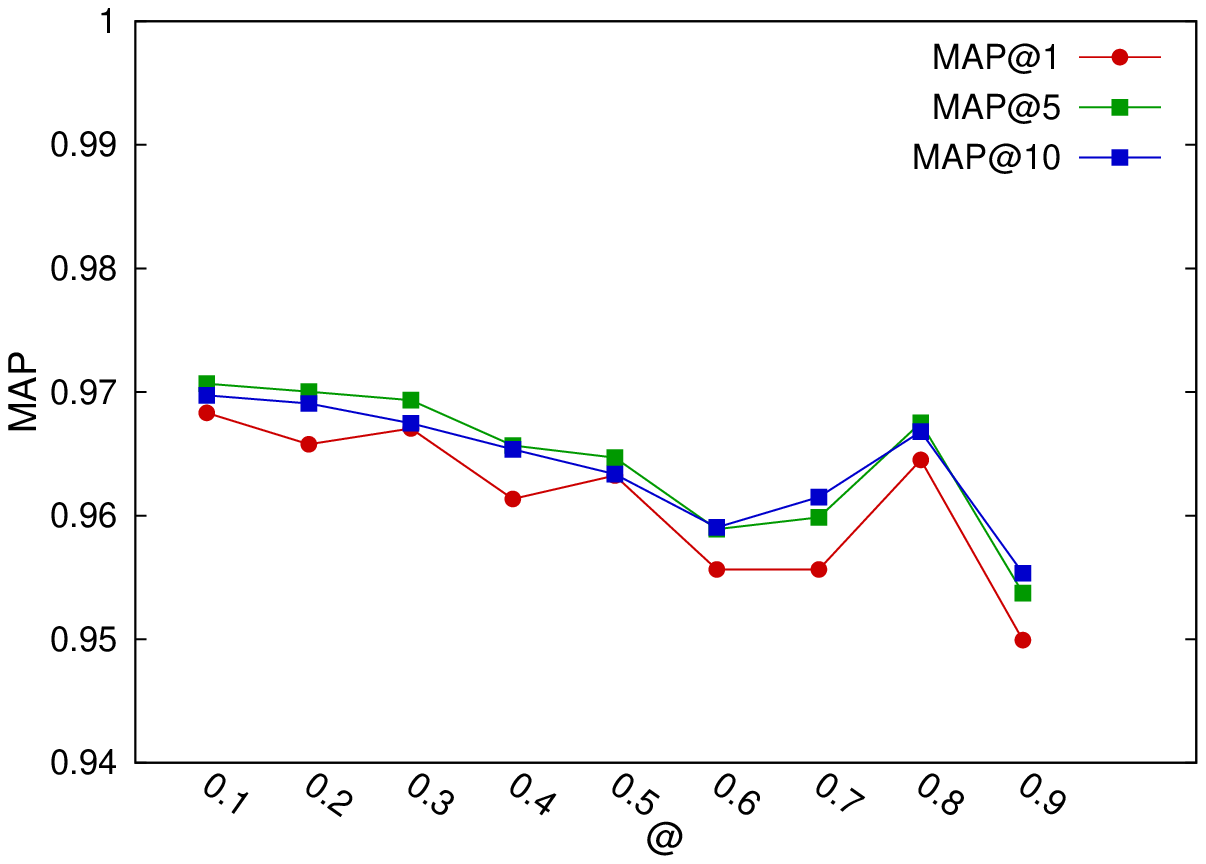}
    }
     \subfloat[{\NetF}]{
    \includegraphics[width=0.4\textwidth]{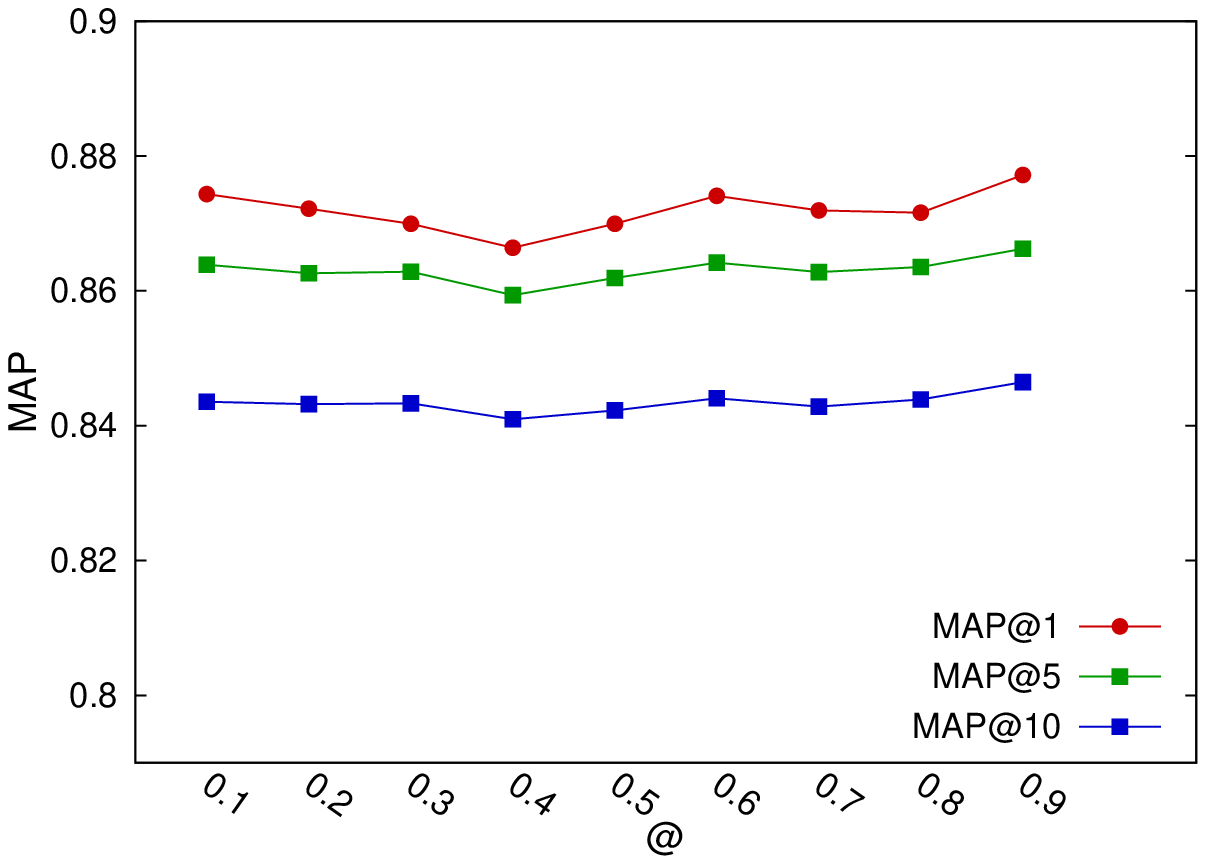}
    }

    \vspace{-1mm}

    \caption{MAP@1, MAP@5, MAP@10 as a function of the value of $\alpha$ for {\ML}-1M, {\ML}-100K and {\kasandr}.}

    \label{fig:alpha_impact}
\end{figure*}
\subsubsection*{Compared baselines}
In order to validate the framework defined in the previous section, we propose to compare the following approaches.

\begin{itemize}
\item {\BPR} \cite{rendle_09} provides an optimization criterion based on implicit feedback; which is the maximum posterior estimator derived from a Bayesian analysis of the pairwise ranking problem, and proposes an algorithm based on Stochastic Gradient Descent to optimize it. The model can further be extended to the explicit feedback case.
\item {\CoFactor} \cite{liang_16}, developed for implicit feedback, constraints the objective of matrix factorization to use jointly item representations with a factorized shifted positive pointwise mutual information matrix of item
co-occurrence counts. The model was found to outperform WMF \cite{Hu:2008} also proposed for implicit feedback.
\item {\LightFM} \cite{kula_15} was first proposed to deal with the problem of cold-start using meta information. As with our approach, it relies on learning the embedding of users and items with the Skip-gram model and optimizes the cross-entropy loss.
\item {\RecNet}$_p$ focuses on the quality of the latent representation of users and items by learning the preference and the representation through the ranking loss $\Loss_p$ (Eq. \ref{eq:embedding_loss}).
\item {\RecNet}$_c$ focuses on the accuracy of the score obtained at the output of the framework and therefore learns the preference and the representation through the ranking loss $\Loss_c$ (Eq. \ref{eq:ranking_loss}).
\item {\RecNet}$_{c,p}$ uses a linear combination of $\Loss_p$ and $\Loss_c$ as the objective function, with $\alpha\in ]0,1[$. We study the two situations presented before (w.r.t. the presence/absence of a supplementary weighting hyper-parameter).
\end{itemize}

\begin{table*}[!htpb]
\centering
\caption{Best parameters for {\RecNet}$_p$, {\RecNet}$_c$ and {\RecNet}$_{c,p}$ when prediction is done on only shown offers; $k$ denotes the dimension of embeddings, $\lambda$ the regularization parameter. We also report the number of hidden units per layer. } \vspace{-2mm}
\label{tab:param}
\resizebox{\textwidth}{!}{\begin{tabular}{|c|ccc|ccc|ccc|ccc|}
\hline
                 & \multicolumn{3}{c|}{\ML-100K} & \multicolumn{3}{c|}{\ML-1M} & \multicolumn{3}{c|}{\NetF}& \multicolumn{3}{c|}{\kasandr}\\ \hline
                 &  {\RecNet}$_c$  & {\RecNet}$_p$&{\RecNet}$_{c,p}$ &{\RecNet}$_c$   &{\RecNet}$_p$ &{\RecNet}$_{c,p}$ &{\RecNet}$_c$   &{\RecNet}$_p$ &{\RecNet}$_{c,p}$&{\RecNet}$_c$   &{\RecNet}$_p$ &{\RecNet}$_{c,p}$  \\ \hline
$k$            &$1$&$2$&$2$&$16$&$1$&$1$&$9$&$2$&$6$ &$19$&$1$&$18$    \\
$\lambda$      &$0.05$&$0.005$&$0.005$&$0.05$&$0.0001$&$0.001$&$0.05$&$0.01$&$0.05$&$0.0001$&$0.05$&$0.005$   \\
\# units &$32$&$64$&$16$&$32$&$16$&$32$&$64$&$16$& $16$ &$64$&$16$&$64$      \\ \hline
\end{tabular}
}
\end{table*}

\begin{table*}[!htpb]
\centering
\caption{Best parameters for {\RecNet}$_p$, {\RecNet}$_c$ and {\RecNet}$_{c,p}$ when prediction is done on all offers; $k$ denotes the dimension of embeddings, $\lambda$ the regularization parameter. We also report the number of hidden units per layer.}\vspace{-2mm}
\label{tab:param_all}
\resizebox{\textwidth}{!}{\begin{tabular}{|c|ccc|ccc|ccc|ccc|}
\hline
                 & \multicolumn{3}{c|}{\ML-100K} & \multicolumn{3}{c|}{\ML-1M} & \multicolumn{3}{c|}{\NetF}& \multicolumn{3}{c|}{\kasandr}\\ \hline
                 &  {\RecNet}$_c$  & {\RecNet}$_p$&{\RecNet}$_{c,p}$ &{\RecNet}$_c$   &{\RecNet}$_p$ &{\RecNet}$_{c,p}$ &{\RecNet}$_c$   &{\RecNet}$_p$ &{\RecNet}$_{c,p}$&{\RecNet}$_c$   &{\RecNet}$_p$ &{\RecNet}$_{c,p}$  \\ \hline
$k$            &$15$&$5$&$8$&$2$&$11$&$2$&$3$&$13$&$1$ &$4$&$16$&$14$    \\
$\lambda$      &$0.001$&$0.001$&$0.001$&$0.05$&$0.0001$&$0.001$&$0.0001$&$0.001$&$0.001$&$0.001$&$0.0001$&$0.05$   \\
\# units &$32$&$16$&$16$&$32$&$64$&$32$&$32$&$64$& $64$ &$32$&$64$&$64$      \\ \hline
\end{tabular}
}
\end{table*}

\subsubsection*{Evaluation protocol}
For each dataset, we sort the interactions according to time, and take 80\% for training the model and the remaining 20\% for testing it. In addition, we remove all users and offers which do not occur during the training phase.
We study two different scenarios for the prediction phase: (1) for a given user, the prediction is done only on the items that were shown to him or her; (2) the prediction is done over the set of all items, regardless of any knowledge about previous interactions. In the context of movie recommendation, a shown item is defined as a movie for which the given user provided a rating. For {\kasandr}, the definition is quite straight-forward as the data were collected from an on-line advertising platform, where the items are displayed to the users, who can either click or ignore them.

\smallskip

The first setting is arguably the most common in academic research, but is abstracted from the real-world problem as at the time of making the recommendation, the notion of shown items is not available, therefore forcing the RS to  consider the set of all items as potential candidates. As a result, in this setting, for \ML-100K, \ML-1M,  \kasandr\ and  \NetF,\ we only consider in average 25, 72, 6 and 8 items for prediction per user. The goal of the second setting is to reflect this real-world scenario, and we can expect lower results than in the first setting as the size of the search space of items increases considerably. To summarize, predicting only among the items that were shown to user evaluates the model's capability of retrieving highly rated items among the shown ones, while predicting among all items measures the performance of the model on the basis of its ability to recommend offers which user would like to engage in. 

\smallskip

All comparisons are done based on a common ranking metric, namely the Mean Average Precision (MAP). First, let us recall that the Average Precision (AP$@\ell$) is defined over the precision, $Pr$ (fraction of recommended items clicked by the user), at rank $\ell$.
$$
\text{AP}@\ell=\frac{1}{\ell}\sum_{j=1}^\ell r_{j} Pr(j),
$$
where the relevance judgments, $r_j$, are binary (i.e. equal to $1$ when the item is clicked or preferred, and 0 otherwise). Then, the mean of these AP's across all users is the MAP. In the following results, we report MAP at different rank $\ell= 1$ and $10$.

\subsubsection*{Hyper-parameters tuning}

First, we provide a detailed study of the impact of the different hyper-parameters involved in the proposed framework $\RecNet_.$. For all datasets, hyper-parameters tuning is done on a separate validation set.
\begin{itemize}
    \item The size of the embedding is chosen among $k \in \{1,\ldots,20\}$. The impact of $k$ on the performance is presented in Figure \ref{fig:emb_dimension}.
    \item We use $\ell_2$ regularization on the embeddings and choose $\lambda\in\{0.0001,0.001,0.005,0.01,0.05\}$.
    \item We run {\RecNet} with 1 hidden layer with relu activation functions, where the number of hidden units is chosen in $\{16,32,64\}$.
    \item In order to train $\RecNet$, we use ADAM \cite{KingmaB14} and found the learning rate $\eta=1e-3$ to be more efficient for all our settings.
    For other parameters involved in Adam, i.e., the exponential decay rates for the moment estimates, we keep the default values ($\beta_1=0.9$, $\beta_{2}=0.999$ and $\epsilon=10^{-8}$).
    \item Finally, we fix the number of epochs to be $T=10,000$ in advance and the size of mini-batches to $n=512$.
   \item One can see that all three versions of $\RecNet$ perform the best with a quite small number of hidden units, only one hidden layer and a low dimension for the representation. As a consequence, they involve a few number of parameters to tune while training.\cmmnt{ and present an interesting computational complexity compared to other state-of-the-art approaches.}
   \item In terms of the ability to recover a relevant ranked list of items for each user, we also tune the hyper-parameter $\alpha$ (Eq. \ref{eq:rankingLoss_alpha}) which balances the weight given to the two terms in $\RecNet_{c,p}$. These results are shown in Figure \ref{fig:alpha_impact}, where the values of $\alpha$ are taken in the interval $[0,1]$. While it seems to play a significant role on {\ML}-100K and {\kasandr}, we can see that for {\ML}-1M the results in terms of MAP are stable, regardless the value of $\alpha$.
\end{itemize}

From Figure \ref{fig:emb_dimension}, when prediction is done on the interacted offers, it is clear that best MAP@1 results are generally obtained with small sizes of item and user embedded vector spaces  $k$.\cmmnt{, which are the same as the size of the feature vector space.} These empirical results support our theoretical analysis where we found that small $k$ induces smaller generalization bounds. This observation on the dimension of embedding is also in agreement with the conclusion of \cite{kula_15}, which uses the same technique for representation learning. For instance, one can see that on {\ML}-1M, the highest MAP is achieved with a dimension of embedding equals to $1$.  Since in the interacted offers setting, the prediction is done among the very few shown offers, \RecNet\ makes non-personalized recommendations. This is due to the fact that \cmmnt{In the case of, {\RecNet}$_p$}having $k=1$ means that the recommendations for a given user with a positive (negative) value is done by sorting the positive (negative) items according to their learned embeddings, and in some sense, can therefore be seen as a bi-polar popularity model. This means that in such cases popularity and non-personalized based approaches are perhaps the best way to make recommendations.  For reproducibility purpose, we report the best combination of parameters for each variant of {\RecNet} in Table \ref{tab:param} and Table \ref{tab:param_all}.

\begin{table*}[!htpb]
\centering
\caption{Results of all state-of-the-art approaches for implicit feedback when prediction is done only on offers shown to users. The best result is in bold, and a $\DA$ indicates a result that is statistically significantly worse than the best, according to a Wilcoxon rank sum test with $p < .01$.}\vspace{-2mm}
\label{tab:results_warm_interacted}
\resizebox{0.9\textwidth}{!}{\begin{tabular}{|c|cc|cc|cc|cc|}
\hline
                   & \multicolumn{2}{c|}{\ML-100K} & \multicolumn{2}{c|}{\ML-1M} & \multicolumn{2}{c|}{\NetF}& \multicolumn{2}{c|}{\kasandr}\\ \hline
            & MAP@1    & MAP@10 &MAP@1   & MAP@10&MAP@1   &MAP@10 & MAP@1   &  MAP@10 \\ \hline
BPR-MF      & $0.613\DA$  &  $0.608\DA$   &$0.788\DA$&$0.748\DA$&$\textbf{0.909}$&$0.842\DA$ &$0.857\DA$&$0.857\DA$     \\
LightFM        &     $0.772\DA$	    &    $0.770\DA$   &$0.832\DA$&$0.795\DA$&$0.800\DA$& $0.793\DA$ &$0.937\DA$&$0.936\DA$  \\
CoFactor          &      $0.718\DA$   &  $0.716\DA$        &$0.783\DA$&$0.741\DA$&$0.693\DA$&$0.705\DA$ &$0.925\DA$&$0.918\DA$\\
{\RecNet}$_c$      & $\textbf{0.894}$    & $\textbf{0.848}$  &$0.877\DA$&$0.835$&$0.880\DA$&$\textbf{0.847}$&$0.958\DA$&$0.963\DA$  \\
{\RecNet}$_p$      &   $0.881\DA$  &   $0.846$  &$0.876\DA$&$\textbf{0.839}$&$0.875\DA$&$0.844$ &$0.915\DA$&$0.923\DA$      \\
{\RecNet}$_{c,p}$     &  $0.888\DA$   & $0.842$       &$\textbf{0.884}$&$\textbf{0.839}$&$0.879\DA$&$\textbf{0.847}$ &$\mathbf{0.970}$&$\textbf{0.973}$   \\ \hline
\end{tabular}
}
\end{table*}
\begin{table*}[!htpb]
\centering
\caption{Results of all state-of-the-art approaches for recommendation on all implicit feedback data sets when prediction is done on all offers. The best result is in bold, and a $\DA$ indicates a result that is statistically significantly worse than the best, according to a Wilcoxon rank sum test with $p < .01$}
\label{tab:results_warm_all}
\resizebox{0.9\textwidth}{!}{\begin{tabular}{|c|cc|cc|cc|cc|}
\hline
                   & \multicolumn{2}{c|}{\ML-100K} & \multicolumn{2}{c|}{\ML-1M} & \multicolumn{2}{c|}{\NetF}& \multicolumn{2}{c|}{\kasandr}\\ \hline
                & MAP@1     & MAP@10 &MAP@1   &  MAP@10&MAP@1   &MAP@10 & MAP@1  & MAP@10 \\ \hline
BPR-MF          &  $0.140\DA$      &  $\textbf{0.261}$      &$0.048\DA$&$0.097\DA$&$0.035\DA$& $0.072\DA$&$0.016\DA$&$0.024\DA$     \\
LightFM         &  $0.144\DA$  & $0.173\DA$     &$0.028\DA$&$0.096\DA$&$0.006\DA$& $0.032\DA$ &$0.002\DA$&$0.003\DA$  \\
CoFactor          &   $ 0.056\DA$     &      $0.031\DA$      &$0.089\DA$&$0.033\DA$&$0.049\DA$&$0.030\DA$&$0.002\DA$&$0.001\DA$\\
{\RecNet}$_c$      &$0.106\DA$       & $0.137 \DA$  &$0.067\DA$&$0.093\DA$&$0.032\DA$&$0.048\DA$ &$0.049\DA$&$0.059\DA$  \\
{\RecNet}$_p$      &  $\textbf{0.239}$   &$0.249$     &$\textbf{0.209}$&$\textbf{0.220}$&$\textbf{0.080}$&$\textbf{0.089}$ &$0.100\DA$&$0.100\DA$     \\
{\RecNet}$_{c,p}$      &   $0.111\DA$ &      $0.134\DA$    &$0.098\DA$&$0.119\DA$&$0.066\DA$&$0.087$ &$\textbf{0.269}$&$\textbf{0.284}$  \\ \hline

\end{tabular}
}
\end{table*}
\subsection{Results}

Hereafter, we compare and summarize the performance of {\RecNet}$_.$ with the baseline methods on various data sets. Empirically, we observed that the version of $\RecNet_{c,p}$ where both $\Loss_c$ and $\Loss_p$ have an equal weight while training gives better results on average, and we decided to only report these results  later.

\medskip

Tables \ref{tab:results_warm_interacted} and \ref{tab:results_warm_all} report all results. In addition, in each case, we statistically compare the performance of each algorithm, and we use bold face to indicate the highest performance, and the symbol $\DA$ indicates that performance is significantly worst than the best result, according to a Wilcoxon rank sum test used at a p-value threshold of $0.01$ \cite{lehmann_06}.


\subsubsection*{Setting 1 : interacted items}
When the prediction is done over offers which user interacted with (Table \ref{tab:results_warm_interacted}), the {\RecNet} architecture, regardless the weight given to $\alpha$, beats all the other algorithms on {\kasandr}, {\ML}-100K and {\ML}-1M. However, on {\NetF}, BPR-MF outperforms our approach in terms of MAP@1. This may be owing to the fact that the binarized \NetF\ movie data set is strongly biased towards the popular movies and usually, the majority of users have watched one or the other popular movies in such data sets and rated them well. In {\NetF}, around $75\%$ of the users have given ratings greater to 4 to the top-10 movies. We believe that this phenomenon adversely affects the performance of {\RecNet}. However, on {\kasandr}, which is the only true implicit dataset {\RecNet} significantly outperforms all other approaches.

\subsubsection*{Setting 2 : all items}
When the prediction is done over all offers (Table \ref{tab:results_warm_all}), we can make two observations. First, all the algorithms encounters an extreme drop of their performance in terms of MAP. Second, {\RecNet} framework significantly outperforms all other algorithms on all datasets, and this difference is all the more important on {\kasandr}, where for instance {\RecNet$_{c,p}$} is in average 15 times more efficient. We believe, that our model is a fresh departure from the models which learn pairwise ranking function without the knowledge of embeddings or which learn embeddings without learning any pairwise ranking function. While learning pairwise ranking function, our model is aware of the learned embeddings so far and vice-versa. We demonstrate that the simultaneous learning of two ranking functions helps in learning hidden features of implicit data and improves the performance of {\RecNet}.

\subsubsection*{Comparison between {\RecNet} versions}
One can note that while optimizing ranking losses by Eq. \ref{eq:rankingLoss} or Eq. \ref{eq:ranking_loss} or Eq. \ref{eq:embedding_loss}, we simultaneously learn representation and preference function; the main difference is the amount of emphasis we put in learning one or another. The results presented in both tables tend to demonstrate that, in almost all cases, optimizing the linear combination of the pairwise-ranking loss and the embedding loss ({\RecNet}$_{c,p}$) indeed increases the quality of overall recommendations than optimizing standalone losses to learn embeddings and pairwise preference function. For instance, when the prediction is done over offers which user interacted with (Table \ref{tab:results_warm_interacted}), ({\RecNet}$_{c,p}$) outperforms ({\RecNet}$_{p}$) and ({\RecNet}$_{c}$) on \ML-1M, \kasandr\ and \NetF. When prediction is done on all offers (Table \ref{tab:results_warm_all}), ({\RecNet}$_{c,p}$) outperforms ({\RecNet}$_{p}$) and ({\RecNet}$_{c}$) on \kasandr. Thus, in case of interacted offers setting, optimizing ranking and embedding loss simultaneously boosts performance on all datasets. However, in the setting of  all offers, optimizing both losses simultaneously is beneficial in case of true implicit feedback datasets such as \kasandr (recall that all other datasets were synthetically made implicit).

\section{Conclusion}\label{sec:conclusion}

We presented and analyzed a learning to rank framework for recommender systems which consists of learning user preferences over items. We showed that the minimization of pairwise ranking loss over user preferences involves dependent random variables and provided a theoretical analysis by proving the consistency of the empirical risk minimization in the worst case where all users choose a minimal number of positive and negative items. From this analysis we then proposed {\RecNet}, a new neural-network based model for learning the user preference, where both the user's and item's representations and the function modeling the user's preference over pairs of items are learned simultaneously. The learning phase is guided using a ranking objective that can capture the ranking ability of the prediction function as well as the expressiveness of the learned embedded space, where the preference of users over items is respected by the dot product function defined over that space. The training of {\RecNet} is carried out using the back-propagation algorithm in mini-batches defined over a user-item matrix containing implicit information in the form of subsets of preferred and non-preferred items. The learning capability of the model over both prediction and representation problems show their interconnection and also that the proposed double ranking objective allows to conjugate them well.  We assessed and validated the proposed approach through extensive experiments, using four popular collections proposed for the task of recommendation. Furthermore, we propose to study two different settings for the prediction phase and demonstrate that the performance of each approach is strongly impacted by the set of items considered for making the prediction.

\medskip

For future work, we would like to extend {\RecNet} in order to take into account additional contextual information regarding users and/or items. More specifically, we are interested in the integration of data of different natures, such as text or demographic information. We believe that this information can be taken into account without much effort and by doing so, it is possible to improve the performance of our approach and tackle the problem of providing recommendation for new users/items at the same time, also known as the cold-start problem. The second important extension will be the development of an on-line version of the proposed algorithm in order to make the approach suitable for real-time applications and on-line advertising. Finally, we have shown that choosing a suitable $\alpha$, which controls the the trade-off between ranking and embedding loss, greatly impact the performance of the proposed framework, and we believe that an interesting extension will be to learn automatically this hyper-parameter, and to make it adaptive during the training phase.

\section*{Acknowledgements}
This work was partly done under the Calypso project supported by the FEDER program from the R\'egion Auvergne-Rh\^one-Alpes. The research of Yury Maximov at LANL was supported by Center of Non Linear Studies (CNLS).

\begin{IEEEbiographynophoto}{Sumit Sidana}
is a PhD student at Grenoble Alpes University, Grenoble. He received a bachelor’s degree in Information Technology from SVIET, India and a master’s degree in computer science from IIIT, Hyderabad. His research interests include machine learning, recommender systems, probabilistic graphical models and deep learning.
\end{IEEEbiographynophoto}
\begin{IEEEbiographynophoto}{Mikhail Trofimov}
is a PhD student at Federal Research Center «Computer Science and Control» of Russian Academy of Sciences, Moscow. He received a bachelor's degree in applied math and physics and master's degrees in intelligent data analysis from Moscow Institute Of Physics and Technology. His research interests include machine learning on sparse data, tensor approximation methods and counterfactual learning.
\end{IEEEbiographynophoto}
\begin{IEEEbiographynophoto}{Oleg Gorodnitskii}
is a MSc student at Skoltech Institute of Science and Technology. He received a bachelor's degree in applied math and physics from Moscow Institute Of Physics and Technology. His research interests include machine learning, recommender systems, optimization methods and deep learning.
\end{IEEEbiographynophoto}
\begin{IEEEbiographynophoto}{Charlotte Laclau} received the PhD degree from the University of Paris Descartes in 2016. She is a a postdoctoral researcher in the Machine Learning group in the University of Grenoble Alpes. Her research interests include statistical machine learning, data mining and particularly unsupervised learning in information retrieval.
\end{IEEEbiographynophoto}
\begin{IEEEbiographynophoto}{Yury Maximov}
is a postdoc at CNLS and the Theoretical division of Los Alamos National Laboratory, and Assistant Professor of the Center of Energy Systems at Skolkovo Institute of Science and Technology. His research is in optimization methods and machine learning theory.
\end{IEEEbiographynophoto}
\begin{IEEEbiographynophoto} {Massih-R\'eza Amini}
 is a Professor in the University of Grenoble Alpes and head of the Machine Learning group. His  research is in statistical machine learning and he has contributed in developing machine learning techniques for information retrieval and text mining.
\end{IEEEbiographynophoto}

%




\end{sloppypar}

\begin{thebibliography}{10}
\providecommand{\url}[1]{#1}
\csname url@samestyle\endcsname
\providecommand{\newblock}{\relax}
\providecommand{\bibinfo}[2]{#2}
\providecommand{\BIBentrySTDinterwordspacing}{\spaceskip=0pt\relax}
\providecommand{\BIBentryALTinterwordstretchfactor}{4}
\providecommand{\BIBentryALTinterwordspacing}{\spaceskip=\fontdimen2\font plus
\BIBentryALTinterwordstretchfactor\fontdimen3\font minus
  \fontdimen4\font\relax}
\providecommand{\BIBforeignlanguage}[2]{{%
\expandafter\ifx\csname l@#1\endcsname\relax
\typeout{** WARNING: IEEEtran.bst: No hyphenation pattern has been}%
\typeout{** loaded for the language `#1'. Using the pattern for}%
\typeout{** the default language instead.}%
\else
\language=\csname l@#1\endcsname
\fi
#2}}
\providecommand{\BIBdecl}{\relax}
\BIBdecl

\bibitem{Ricci:2010:RSH:1941884}
F.~Ricci, L.~Rokach, B.~Shapira, and P.~B. Kantor, \emph{Recommender Systems
  Handbook}, 1st~ed.\hskip 1em plus 0.5em minus 0.4em\relax New York, NY, USA:
  Springer-Verlag New York, Inc., 2010.

\bibitem{reference/rsh/LopsGS11}
P.~Lops, M.~de~Gemmis, and G.~Semeraro, ``Content-based recommender systems:
  State of the art and trends.'' in \emph{Recommender Systems Handbook},
  F.~Ricci, L.~Rokach, B.~Shapira, and P.~B. Kantor, Eds.\hskip 1em plus 0.5em
  minus 0.4em\relax Springer, 2011, pp. 73--105.

\bibitem{ir2004010}
R.~White, J.~M. Jose, and I.~Ruthven, ``Comparing explicit and implicit
  feedback techniques for web retrieval: {TREC-10} interactive track report,''
  in \emph{Proceedings of {TREC}}, 2001.

\bibitem{DBLP:conf/kdd/WangWY15}
H.~Wang, N.~Wang, and D.~Yeung, ``Collaborative deep learning for recommender
  systems,'' in \emph{Proceedings of {SIGKDD}}, 2015, pp. 1235--1244.

\bibitem{Usunier:1121}
N.~Usunier, M.~Amini, and P.~Gallinari, ``A data-dependent generalisation error
  bound for the {AUC},'' in \emph{ICML'05 workshop on ROC Analysis in Machine
  Learning}, Bonn, Germany, 2005.

\bibitem{Amini:15}
M.-R. Amini and N.~Usunier, \emph{Learning with Partially Labeled and
  Interdependent Data}.\hskip 1em plus 0.5em minus 0.4em\relax New York, NY,
  USA: Springer, 2015.

\bibitem{Janson04RSA}
S.~Janson, ``{Large Deviations for Sums of Partly Dependent Random
  Variables},'' \emph{Random Structures and Algorithms}, vol.~24, no.~3, pp.
  234--248, 2004.

\bibitem{UsunierAG05}
N.~Usunier, M.-R. Amini, and P.~Gallinari, ``Generalization error bounds for
  classifiers trained with interdependent data,'' in \emph{Proceedings of
  NIPS}, 2006, pp. 1369--1376.

\bibitem{mcdiarmid89method}
C.~McDiarmid, ``On the method of bounded differences,'' \emph{Survey in
  Combinatorics}, pp. 148--188, 1989.

\bibitem{RalaiAmin15}
L.~Ralaivola and M.~Amini, ``Entropy-based concentration inequalities for
  dependent variables,'' 2015, pp. 2436--2444.

\bibitem{vapnik2000nature}
V.~Vapnik, \emph{The nature of statistical learning theory}.\hskip 1em plus
  0.5em minus 0.4em\relax Springer Science \& Business Media, 2000.

\bibitem{Volkovs:2015}
M.~Volkovs and G.~W. Yu, ``Effective latent models for binary feedback in
  recommender systems,'' in \emph{Proceedings of SIGIR}, 2015, pp. 313--322.

\bibitem{rendle_09}
S.~Rendle, C.~Freudenthaler, Z.~Gantner, and L.~Schmidt{-}Thieme, ``{BPR:}
  bayesian personalized ranking from implicit feedback,'' in \emph{Proceedings
  of {UAI}}, 2009, pp. 452--461.

\bibitem{Leon2012}
L.~Bottou, ``Stochastic gradient descent tricks,'' in \emph{Neural Networks:
  Tricks of the Trade - Second Edition}, 2012, pp. 421--436.

\bibitem{Ailon08anefficient}
N.~Ailon and M.~Mohri, ``An efficient reduction of ranking to classification,''
  in \emph{Proceedings of {COLT}}, (2008), pp. 87--98.

\bibitem{word_emb}
T.~Mikolov, K.~Chen, G.~Corrado, and J.~Dean, ``Efficient estimation of word
  representations in vector space,'' \emph{CoRR}, vol. abs/1301.3781, 2013.

\bibitem{mikolov_13}
T.~Mikolov, I.~Sutskever, K.~Chen, G.~S. Corrado, and J.~Dean, ``Distributed
  representations of words and phrases and their compositionality,'' in
  \emph{Proceedings of NIPS}, 2013, pp. 3111--3119.

\bibitem{levy_14}
O.~Levy and Y.~Goldberg, ``Neural word embedding as implicit matrix
  factorization,'' in \emph{Proceedings of NIPS}, 2014, pp. 2177--2185.

\bibitem{guardia_15}
{\'{E}}.~Gu{\`{a}}rdia{-}Sebaoun, V.~Guigue, and P.~Gallinari, ``Latent
  trajectory modeling: {A} light and efficient way to introduce time in
  recommender systems,'' in \emph{Proceedings of RecSys}, 2015, pp. 281--284.

\bibitem{liang_16}
D.~Liang, J.~Altosaar, L.~Charlin, and D.~M. Blei, ``Factorization meets the
  item embedding: Regularizing matrix factorization with item co-occurrence,''
  in \emph{Proceedings of RecSys}, 2016, pp. 59--66.

\bibitem{DBLP:conf/kdd/GrbovicRDBSBS15}
M.~Grbovic, V.~Radosavljevic, N.~Djuric, N.~Bhamidipati, J.~Savla, V.~Bhagwan,
  and D.~Sharp, ``E-commerce in your inbox: Product recommendations at scale,''
  in \emph{Proceedings of {SIGKDD}}, 2015, pp. 1809--1818.

\bibitem{He:2017:NCF:3038912.3052569}
X.~He, L.~Liao, H.~Zhang, L.~Nie, X.~Hu, and T.~Chua, ``Neural collaborative
  filtering,'' in \emph{Proceedings of {WWW}}, 2017, pp. 173--182.

\bibitem{vasile_16}
F.~Vasile, E.~Smirnova, and A.~Conneau, ``Meta-prod2vec: Product embeddings
  using side-information for recommendation,'' in \emph{Proceedings of RecSys},
  2016, pp. 225--232.

\bibitem{Liu:2009}
T.~Liu, ``Learning to rank for information retrieval,'' \emph{Foundations and
  Trends in Information Retrieval}, vol.~3, no.~3, pp. 225--331, 2009.

\bibitem{Crammer01}
K.~Crammer and Y.~Singer, ``Pranking with ranking,'' in \emph{Proceedings of
  NIPS}, 2001, pp. 641--647.

\bibitem{Li08}
P.~Li, C.~J.~C. Burges, and Q.~Wu, ``Mcrank: Learning to rank using multiple
  classification and gradient boosting,'' in \emph{Proceedings of NIPS}, 2007,
  pp. 897--904.

\bibitem{Shi:2010}
Y.~Shi, M.~Larson, and A.~Hanjalic, ``List-wise learning to rank with matrix
  factorization for collaborative filtering,'' in \emph{Proceedings of RecSys},
  2010, pp. 269--272.

\bibitem{Xu07}
J.~Xu and H.~Li, ``Adarank: a boosting algorithm for information retrieval,''
  in \emph{Proceedings of {SIGIR}}, 2007, pp. 391--398.

\bibitem{Xu08}
J.~Xu, T.~Liu, M.~Lu, H.~Li, and W.~Ma, ``Directly optimizing evaluation
  measures in learning to rank,'' in \emph{Proceedings of {SIGIR}}, 2008, pp.
  107--114.

\bibitem{Cohen99}
W.~W. Cohen, R.~E. Schapire, and Y.~Singer, ``Learning to order things,''
  \emph{J. Artif. Intell. Res. {(JAIR)}}, vol.~10, pp. 243--270, 1999.

\bibitem{Freund03}
Y.~Freund, R.~D. Iyer, R.~E. Schapire, and Y.~Singer, ``An efficient boosting
  algorithm for combining preferences,'' \emph{Journal of Machine Learning
  Research}, vol.~4, pp. 933--969, 2003.

\bibitem{Joachims02}
T.~Joachims, ``Optimizing search engines using clickthrough data,'' in
  \emph{Proceedings of {SIGKDD}}, 2002, pp. 133--142.

\bibitem{PessiotTUAG07}
J.~Pessiot, T.~Truong, N.~Usunier, M.~Amini, and P.~Gallinari, ``Learning to
  rank for collaborative filtering,'' in \emph{Proceedings of {ICEIS}}, 2007,
  pp. 145--151.

\bibitem{Caruana:1995}
R.~Caruana, S.~Baluja, and T.~M. Mitchell, ``Using the future to sort out the
  present: Rankprop and multitask learning for medical risk evaluation,'' in
  \emph{Proceedings of NIPS}, 1995, pp. 959--965.

\bibitem{DBLP:conf/icml/BurgesSRLDHH05}
C.~J.~C. Burges, T.~Shaked, E.~Renshaw, A.~Lazier, M.~Deeds, N.~Hamilton, and
  G.~N. Hullender, ``Learning to rank using gradient descent,'' in
  \emph{Proceedings of ICML}, 2005, pp. 89--96.

\bibitem{DBLP:conf/icann/RigutiniPMB08}
L.~Rigutini, T.~Papini, M.~Maggini, and M.~Bianchini, ``A neural network
  approach for learning object ranking,'' in \emph{Proceedings of {ICANN}},
  2008, pp. 899--908.

\bibitem{DBLP:journals/tnn/RigutiniPMS11}
L.~Rigutini, T.~Papini, M.~Maggini, and F.~Scarselli, ``Sortnet: Learning to
  rank by a neural preference function,'' \emph{{IEEE} Trans. Neural Networks},
  vol.~22, no.~9, pp. 1368--1380, 2011.

\bibitem{Harper:2015:MDH:2866565.2827872}
F.~M. Harper and J.~A. Konstan, ``The movielens datasets: History and
  context,'' \emph{ACM Trans. Interact. Intell. Syst.}, vol.~5, no.~4, pp.
  19:1--19:19, Dec. 2015.

\bibitem{DBLP:conf/sigir/sidana17}
S.~Sidana, C.~Laclau, M.-R. Amini, G.~Vandelle, and A.~Bois-Crettez, ``Kasandr:
  A large-scale dataset with implicit feedback for recommendation,'' in
  \emph{Proceedings of {SIGIR}}, 2017.

\bibitem{Hu:2008}
Y.~Hu, Y.~Koren, and C.~Volinsky, ``Collaborative filtering for implicit
  feedback datasets,'' in \emph{Proceedings of {ICDM}}, 2008, pp. 263--272.

\bibitem{kula_15}
M.~Kula, ``Metadata embeddings for user and item cold-start recommendations,''
  in \emph{Proceedings of the 2nd Workshop on New Trends on Content-Based
  Recommender Systems co-located with RecSys.}, 2015, pp. 14--21.

\bibitem{KingmaB14}
D.~P. Kingma and J.~Ba, ``Adam: {A} method for stochastic optimization,''
  \emph{CoRR}, vol. abs/1412.6980, 2014.

\bibitem{lehmann_06}
E.~Lehmann and H.~D'Abrera, \emph{Nonparametrics: statistical methods based on
  ranks}.\hskip 1em plus 0.5em minus 0.4em\relax Springer, 2006.

\end{thebibliography}
\end{document}